\pdfoutput=1

\documentclass[11pt]{article}

\newif\ifshowauthors
\showauthorstrue

\ifshowauthors
    \usepackage{acl}
\else 
    \usepackage[review]{acl}
\fi

\usepackage{times}
\usepackage{latexsym}

\usepackage[T1]{fontenc}

\usepackage[utf8]{inputenc}

\usepackage{microtype}

\usepackage{inconsolata}

\usepackage{package}
\newcommand{\ourdata}{\textsc{CardBench}\xspace}
\newcommand{\ourmethod}{\textsc{CardGen}\xspace}

\definecolor{cornellred}{rgb}{0.7, 0.11, 0.11}
\newcommand{\rred}[1]{{\color{cornellred}#1}}

\definecolor{dartmouthgreen}{rgb}{0.05, 0.5, 0.06}
\newcommand{\green}[1]{{\color{dartmouthgreen}#1}}

\newif\ifreview
\reviewfalse

\title{
Automatic Generation of Model and Data Cards:\\
A Step Towards Responsible AI}

\author{Jiarui Liu \\
  CMU \\
  \texttt{jiaruil5@andrew.cmu.edu} \\ \And
  Wenkai Li \\
  CMU \\
  \texttt{wenkail@andrew.cmu.edu} \\ \AND
  Zhijing Jin \\
  MPI \& ETH Zürich \\
  \texttt{jinzhi@ethz.ch} \\\And
  Mona Diab \\
  CMU \\
  {\texttt{mdiab@andrew.cmu.edu}} \\\\
}

\begin{document}
\maketitle
\begin{abstract}
In an era of model and data proliferation in machine learning/AI especially marked by the rapid advancement of open-sourced technologies, there arises a critical need for standardized consistent documentation. Our work addresses the information incompleteness in current human-generated model and data cards. We propose an automated generation approach using Large Language Models (LLMs). Our key contributions include the establishment of \ourdata, a comprehensive dataset aggregated from over 4.8k model cards and 1.4k data cards, coupled with the development of the \ourmethod pipeline comprising a two-step retrieval process. Our approach exhibits enhanced completeness, objectivity, and faithfulness in generated model and data cards, a significant step in responsible AI documentation practices ensuring better accountability and traceability.\footnote{
\ifreview
\else
    Our code and data is available at \url{https://github.com/jiarui-liu/AutomatedModelCardGeneration}.
\fi
}
\end{abstract}

\section{Introduction}

The landscape of artificial intelligence (AI) has undergone a profound transformation with the recent surge in open-sourced models \citep{villalobos2022machine, sevilla2022compute} and datasets \citep{northcutt2021pervasive, sevilla2022compute}. The trend has been significantly accelerated by the advent of disruptive technologies such as transformers \citep{gruetzemacher2022transformative, vaswani2017attention}. Since this proliferation of accessible models and datasets can have their applications significantly influence various aspects of society, it becomes increasingly important to underscore the necessity for standardized consistent documentation to communicate their performance characteristics accurately \citep{liang2022advances}.

\begin{figure}[t]
    \centering
    \includegraphics[width=\columnwidth]{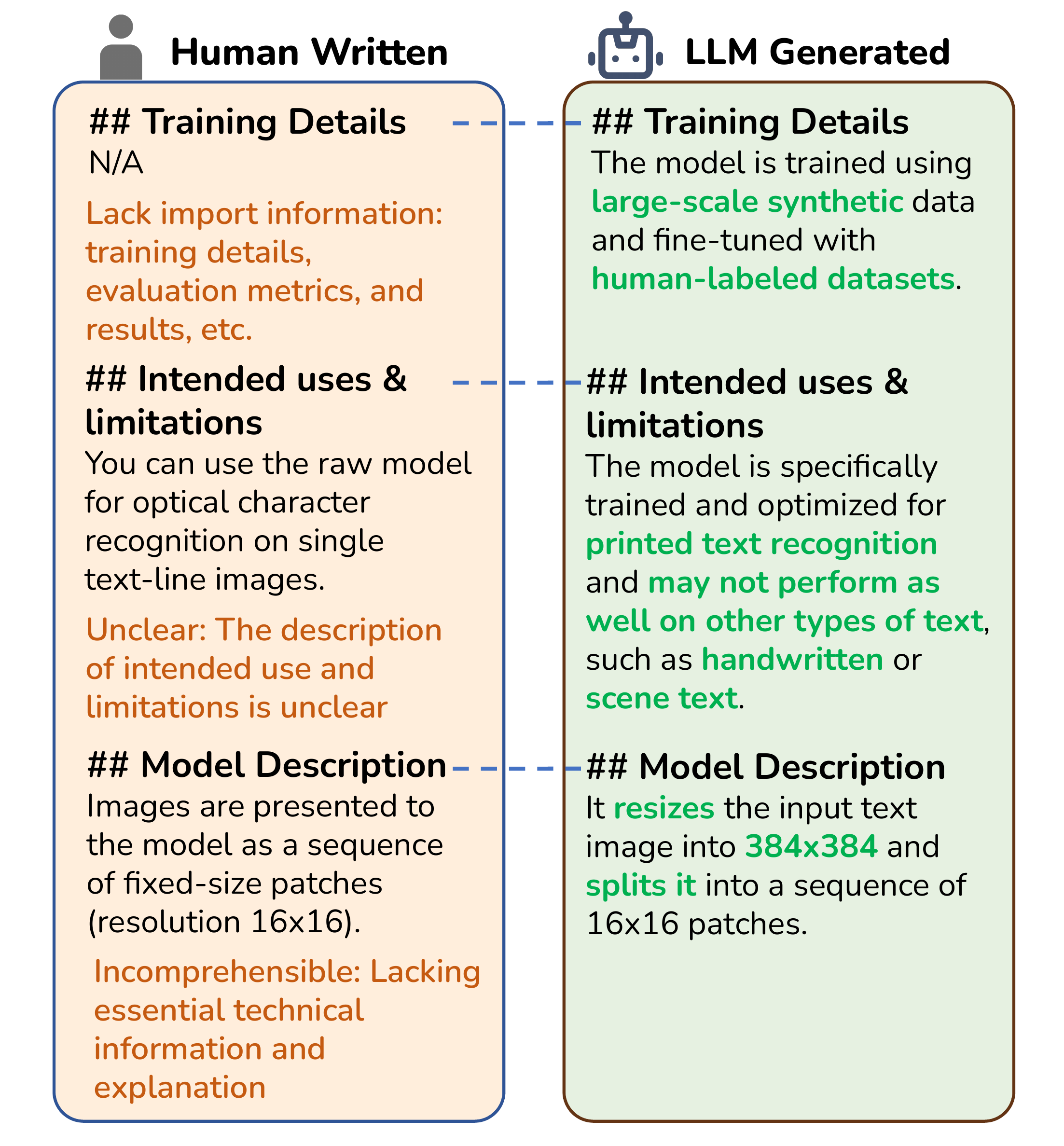}
    \caption{Common problems with manually generated model cards and data cards.}
    \label{fig:problem}
\end{figure}

In this context, model cards proposed by \citet{mitchell2019model} and data cards proposed by \citet{pushkarna2022data}, emerge as necessary documentation tools. These cards bridge the communication gap between model/data creators and product developers, thereby ensuring a comprehensive understanding of the model's/data's capabilities and limitations in both academic and industrial applications \citep{pushkarna2022data, sevilla2022compute, vaswani2017attention, sevilla2022compute}. 
Model/data cards are instrumental in research, offering detailed insights such as data characteristics, sources, etc, as well as model architecture, training procedures, and potential biases and limitations, which accelerates development and reduces error propagation in subsequent models \citep{swayamdipta-etal-2020-dataset}.


\begin{figure*}[t]
    \centering
    \includegraphics[width=\textwidth]{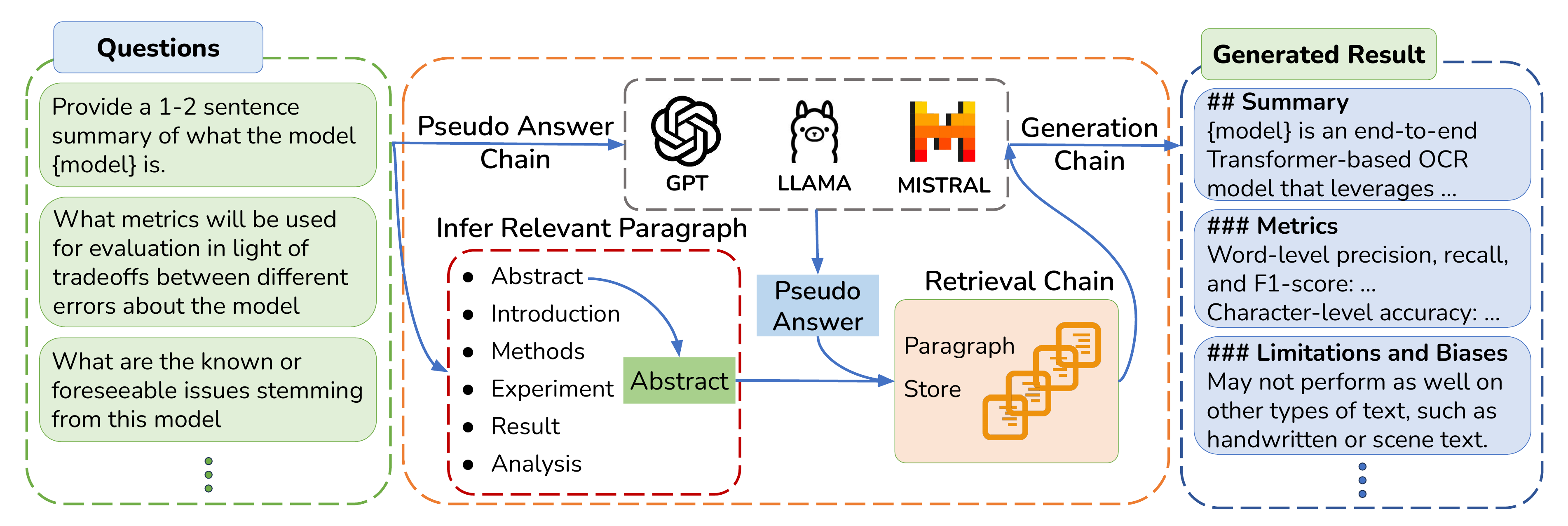}
    \caption{Overview of the \ourmethod pipeline to generate a full model card or a full data card.}
    \label{fig:overview}
\end{figure*}

Inspired by these concepts, HuggingFace (HF) developed card specifications for models and datasets hosted on its website. Despite the release of some available tools to assist model card writing\footnote{\url{https://huggingface.co/spaces/huggingface/Model_Cards_Writing_Tool}}, HF leaves the decision of what to report up to developers. This raises several problems: First, this approach relies heavily on the developers' understanding and interpretation of what should be reported, leading to inconsistencies and potential omissions of critical information \citep{shukla2021model}. Second, there is a tendency among card creators to use existing cards as templates rather than starting from the standardized template provided \citep{pushkarna2022data}. Such variability  compromises the comprehensiveness and reliability of the cards.

With the power of state-of-the-art LLMs \citep{touvron2023llama, brown2020language, ouyang2022training, jiang2023mistral, touvron2023llama}, the automatic generation of model and data cards presents a method to ensure uniformity, consistency, and thoroughness across various model/dataset documentation. To this end, we contribute the following: (1) A novel pioneering initiative to systematically utilize LLMs for automatically generating model/data cards; (2) \ourdata, a curated dataset that encompasses all the associated papers and GitHub READMEs referenced in 4.8k model cards and 1.4k data cards; (3) A novel approach that decomposes the card generation task into multiple sub-tasks, proposing a \ourmethod pipeline including a two-step retrieval process; (4) A novel set of quantitative and qualitative evaluation metrics. We demonstrate that using our pipeline with GPT3.5, we achieve higher scores than human generated cards on completeness, objectivity, and understandability, demonstrating the effectiveness of the \ourmethod pipeline.




\section{Related Work}

\subsection{Accountability and Traceability for AI Systems Through Documentation}
The increasing complexity of AI systems has raised significant concerns regarding their potential biases and lack of transparency, which in turn poses negative implications for users and society \citep{jacovi2021formalizing, Barocas2016BigDD, panch2019artificial, daneshjou2021lack, huang2023survey}. This has motivated the emergence of various documentation frameworks for ML models and datasets:

\paragraph{Model Cards}
\citet{mitchell2019model} introduced the concept of model cards as a framework for the transparent documentation of machine learning (ML) models and provided detailed evaluations across diverse demographic groups and conditions. Subsequent advancements in model card design have included advocating for the generation of consumer labels for ML models \citep{seifert2019towards}, introducing principles for explainable models \citep{phillips2020four}, suggesting other cards as complements to model cards \citep{adkins2022prescriptive, shen2021value}, environmental and financial impact considerations \citep{strubell2019energy}, and some toolkits that help to track and report specific information in ML models \citep{arya2019one, shukla2021model}. 

\paragraph{Data Cards}
In the domain of ML dataset documentation, \citet{gebru2021datasheets} pioneered the concept of datasheets for datasets, followed by the introduction of data statements for NLP data \citep{bender-friedman-2018-data, Bender_Friedman_McMillan-Major_2021}, and the concept of data nutrition labels to aid in better decision-making \citep{holland2020dataset}. \citet{mcmillan2021reusable, hutchinson2021towards} provided comprehensive data card templates. \citet{pushkarna2022data} proposed data cards for responsible AI development. \citet{diaz2022crowdworksheets} introduced CrowdWorkSheets for the transparent documentation of crowdsourced data. Our work builds upon the existing model and data card documentation templates released by HF.

\subsection{Knowledge-Enhanced Text Generation}
LLMs can be augmented with external knowledge sources to improve their reasoning capabilities \citep{lewis2020retrieval, li2022survey}. Retriever, generator, and evaluator are the key components in a standard RAG system. With the advancement of powerful pretrained seq2seq models as generators, numerous studies have concentrated on the evaluation performance:


\vspace{-0.5em}
\paragraph{RAG Text Generation Evaluation}
Due to variations in retrieved content, customized generation pipelines, and user intentions, evaluating the effectiveness of LLM generated texts in a Retrieval-Augmented Generation (RAG) system becomes challenging \citep{huang2023survey, mialon2023augmented}. Traditional $n$-gram based metrics like BLEU \citep{papineni2002bleu}, ROUGE \citep{lin-2004-rouge}, and PARENT-T \citep{wang-etal-2020-towards} are used for assessing the overlap between generated texts and references, but cannot fully grasp the quality nuances of human expectations \citep{honovich-etal-2021-q2, maynez-etal-2020-faithfulness}. Some model-based metrics have later been invented to better align with human judgments without requiring supervision, such as BERTScore \citep{zhang2019bertscore}, MoverScore \citep{zhao-etal-2019-moverscore}, and BARTScore \citep{yuan2021bartscore}. Research has primarily focused on factuality \citep{gou2023critic, chen2023complex, boris2023truth, min2023factscore}, and faithfulness \citep{barrantes2020adversarial, fabbri-etal-2022-qafacteval, santhanam2021rome, laban2023llms, durmus-etal-2020-feqa} of generatd content. Some frameworks have been designed to automate the assessment pipeline utilizing the capabilities of LLMs \citep{es2023ragas, Pietsch_Tanay_Branden_Timo_Bogdan_2020, liu2023gpteval, fu2023gptscore, manakul2023selfcheckgpt}. In this work, we present a comprehensive evaluation of our approach using both traditional metrics and LLM-based automatic metrics. Additionally, we offer a detailed human evaluation of multiple performance aspects, including faithfulness.

\section{Defining the Model/Data Card Generation Task}

\subsection{Task Formulation} 
Denote our test set as $\bm{D}:=\{ (\bm{m}_i, \bm{p}_i, \bm{g}_i)\}_{i=1}^N$ consisting of $N$ triples, each with a human-generated model card $\bm{m}_i$, a direct paper document $\bm{p}_i$, and a direct GitHub README document $\bm{g}_i$. For each question $\bm{q}_j$ from the question template set $\bm{Q}:=\{\bm{q}_j\}_{j=1}^M$, we define a two-stage retrieve-and-generate task $f_1$ and $f_2$. 

The retrieval task $f_1: \mathcal{P} \times \mathcal{G} \times \mathcal{Q} \rightarrow \mathcal{R}$ maps source paper and GitHub documents according to the question to a set of retrieved chunks $\bm{R}$.

The generation task $f_2: \mathcal{R} \times \mathcal{Q} \rightarrow \mathcal{A}$ maps the retrieved chunk set and questions to a space $\mathcal{A}$ that contains generated answers for all questions.

\begin{table*}[t!]
    \centering \small
    \begin{tabularx}{\textwidth}{llX}
    \toprule
    \textbf{Question} & \textbf{Role} & \textbf{Prompt} \\
    \hline
    Summary & Project organizer & Provide a 1-2 sentence summary of what the model is. \\
    \hline
    Description & Project organizer & Provide basic details about the model. This includes the model architecture, training procedures, parameters, and important disclaimers. \\
    \hline
    Direct use & Project organizer & Explain how the model can be used without fine-tuning, post-processing, or plugging into a pipeline. Provide a code snippet if necessary. \\
    \hline
    Bias, risks, limitations & Practical Ethicist & What are the known or foreseeable issues stemming from this model? These include foreseeable harms, misunderstandings, and technical and sociotechnical limitations. \\
    \hline
    Results summary & Developer & Summarize the model evaluation results. \\
    \bottomrule
    \end{tabularx}
    \caption{Template of the most important questions for each section. ``Roles'' are provided as role specifications in Appendix \cref{fig:generation_prompt} for LLMs, and ``prompts'' are provided as queries.}
    \label{tab:question_template}
\end{table*}

\subsection{Structured Generation}
Inspired by the model card design from \citet{mitchell2019model}, HF provides its guidelines about how to fully fill out a model card.\footnote{\url{https://huggingface.co/docs/hub/model-card-annotated}} It suggests a detailed disclosure of the model features and limitations in a published model card. Following the guidelines, we define seven sections including 31 individual questions for generating a complete model card. These sections are model summary, model details, uses, bias and risks, training details, evaluation, and additional information about the proposed model. We have made our full question template for both model cards and data cards accessible in \cref{appn:question_template}. \cref{tab:question_template} highlights the most important questions for each section of the full template.

\section{\ourdata Dataset}

\ourdata contains 4,829 human-generated model cards and 328 data cards with paper and GitHub references. 



\subsection{Dataset Collection}
\paragraph{Data Source and Preprocessing} We identify the model page\footnote{\url{https://huggingface.co/models}} and the dataset page\footnote{\url{https://huggingface.co/datasets}} on HF as data sources. We crawl the model cards and data cards (READMEs) associated with the 10,000 most downloaded models and datasets, respectively, from the HF page as of October 1, 2023. For each collected model card, we use regular expressions to find all valid paper URLs and GitHub repository URLs. We leverage the SciPDF Parser\footnote{\url{https://github.com/titipata/scipdf_parser}} to parse downloaded paper PDFs into a JSON formatted data structure, capturing the paper sections. We further use the GitHub REST API\footnote{\url{https://docs.github.com/en/rest?api}} to obtain README files from each repository. For each collected data card, we devise regular expressions to locate all data cards with the ``Dataset Description'' section, which should contain information such as the dataset homepage, paper link, and GitHub repository. Then, based on the information obtained from the data card, we retrieve and process paper documents and GitHub READMEs as done for model cards.


\vspace{-0.5em}
\paragraph{Evaluation Set Construction} In the absence of standardized and strict content requirements by HF, collected model cards are mostly incomplete, and some examples are even minimally modified copies of existing ones. This variability undermines the reliability of our comparative evaluation against human-generated model cards as a reference metric. In an attempt to mitigate this shortcoming, we curate the highest quality human generated model cards to serve as our evaluation data set. This set comprised a selection of 350 examples that are rewritten by the HF team with their unique disclaimers. Also, for data cards, the majority of those collected are incomplete and lack content readability. In order to have a sufficient number of evaluation sets, we first selected all the data cards with a ``Dataset Description'' section. We then wrote markdown matching logic to obtain 300 examples as our evaluation set based on the word count and the number of sections in the data cards. See \cref{appn:data_collect} for more details on data collection. 

\begin{table}[t]
    \centering
    \resizebox{\columnwidth}{!}{
    \begin{tabular}{lccccc}
    \toprule
    & \multirow{2}{*}{Split} & \multicolumn{2}{c}{Paper} & \multicolumn{2}{c}{GitHub} \\
    \cline{3-6}
    &  & \# Sections & \# Words & \# Sections & \# Words \\
    \hline
    \multirow{2}{*}{ModelCard} & all & 29 & 6810 & 22 & 2495 \\
                                & test& 30 & 6674 & 17 & 1855 \\
    \hline
    \multirow{2}{*}{DataCard} & all & 25 & 5741 & 9 & 975 \\
                                & test& 25 & 5784 & 8 & 816 \\
    \bottomrule
    \end{tabular}
    }
    \caption{Statistics for direct paper documents and repository READMEs for crawled model cards and data cards, in terms of the average number of sections and the average number of words of documents.}
    \label{tab:stat_doc}
\end{table}

\subsection{Data Annotation}
In our methodology for generating model cards, we predominantly focus on the model's design detail itself rather than referencing external methodologies cited in human-generated model cards. It necessitates the identification of the primary paper proposing the model, along with the direct repository reflecting model implementation. The evaluation set is annotated by two ML Master's student researchers who know HF models well and are proficient in English. The process resulted in 294 evaluation examples having both direct paper and repository links. Additionally, to annotate the whole dataset, we prompt \texttt{GPT-3.5-Turbo} \citep{brown2020language} to validate direct source document links, given the context wherein each URL is situated in the model card. We finally obtained 4,829 non-empty ones with either direct paper links or repository links. GPT's annotation reached 98.01\% accuracy according to human validation results on the test set. For data cards, their primary paper link and direct repository responsible for the dataset is within the ``Dataset Description'' section. We finally obtained 865 data cards with either direct paper links or repository links. This gain resulted in 99.7\% accuracy according to human validation results on the 300 data cards test set. See \cref{appn:data_annotate} for human annotation guidelines and prompts for GPT validation.

\subsection{Data Statistics}

We show the overall statistics in \cref{tab:stat_doc} and Appendix \cref{tab:stat_card}. We can observe that our test set, consisting of the set of model cards rewritten by the HF team, are more concise than other developer-written ones. Their corresponding source documents have similar sizes in terms of the number of sections and words.

To explore whether our test set represents the whole dataset well, we look into some model card features obtained with the HF API. Appendix \cref{fig:stats_downloads} shows that test set examples are nearly uniformly distributed compared to the overall dataset in terms of the number of downloads, and task distributions of models/datasets. A comparison of the test set to the whole set is shown in Appendix \cref{fig:task_taxonomy}. See \cref{appn:model_card_task_taxonomy} for additional dataset analyses.




\section{Method: the \ourmethod Pipeline}

\subsection{Overview}
\cref{fig:overview} shows our \ourmethod pipeline. For each $\bm{q}_j$ in $\bm{Q}$, we first prompt LLMs to split $\bm{q}_j$ into a sub-question set. Next, we use LLMs to infer relevant sections as potential knowledge sources, and generate pseudo answers for each sub-question leveraging LLM's own knowledge \citep{hyde}. The pseudo answer is used as a query to get the set $\bm{R}$ of relevant document chunks. We use an LLM to generate answers for the question prepended with highest-ranked document chunks. 

\subsection{Designing the Retriever}
As the process of supervised retrieval necessitates the acquisition of additional crowd-sourced annotations for establishing ground truth sentences for each query, it constitutes a substantial amount of labor. Consequently, we choose to modify the standard RAG retrieval baselines \citep{lewis2020retrieval}, where source documents are ranked based on the inner product similarity with a query question. We develop a two-step retrieval method to improve the retrieval precision: (1) Given all section names of a model's paper and README documents, we prompt the LLM to infer the top-k most plausibly relevant sections. (2) We query the pseudo answer from chunks in the inferred section contents after feeding it into an embedding model. We use the embedding model \texttt{jina-embeddings-v2-base-en} developed by \citet{günther2023jina}. This choice is further verified in \cref{sec:baseline_result,par:embed}. 

\begin{table*}[t]
    \centering \small
    \begin{tabularx}{\textwidth}{llX}
    \toprule
    Metric & Input & Description \\
    \hline
    Factual consistency & $\bm{R}, \bm{A}$ & How much the generated answer is supported by retrieved contexts. \\
    \hline
    Faithfulness & $\bm{Q}, \bm{R}, \bm{A}$ & How much the statements created from the question-answer pair are supported by the retrieved context. \\
    Answer relevance & $\bm{Q}, \bm{A}$ & Relevance score of the answer according to the given question. \\
    Context precision & $\bm{Q}, \bm{R}$ & How much the given context is useful in answering the question. \\
    Context relevance & $\bm{Q}, \bm{R}$ & Whether the question can be answered by relevant sentences extracted from the given context. \\
    \bottomrule
    \end{tabularx}
    \caption{Illustration of the input, along with a description of standard metrics and GPT-based metrics being used. Here $\bm{Q}, \bm{R}, \bm{A}$ represent the questions, retrieved texts, and generated texts, respectively.}
    \label{tab:gpt_metric}
\end{table*}

\subsection{Designing the Generator}
For our \ourmethod pipeline, we test \texttt{Claude 3 Opus} \citep{anthropic2024claude3}, \texttt{GPT-4-Turbo} \citep{openai2023gpt4}, \texttt{GPT-3.5-Turbo} \citep{brown2020language}, \texttt{Llama2 70B Chat} \citep{touvron2023llama}, \texttt{Vicuna 13B V1.5} \citep{zheng2024judging}, \texttt{Llama2 7B Chat} \citep{touvron2023llama}, \texttt{Mistral 7B Instruct} \citep{jiang2023mistral} as backbone LLMs. We generate an answer $\bm{t}_j$ for each question $\bm{q}_j$ based on $\bm{R}$, and concatenate all answers in sequence to form the final model card. To leverage the LLM's strengths in effectively responding to varied questions, we assign specific roles to the LLM tailored to different questions, and outline its expected areas of expertise. The pre-defined roles, such as project organizer, sociotechnical practical ethicist, and developer, are outlined in \cref{tab:question_template,appn:question_template}, as noted by \citet{annotate-model-card}. See \cref{appn:generator} for LLM inference details.

\section{Baselines}
\label{sec:baseline}

We evaluate our two-step retrieval and generation processes in \ourmethod against two baselines:

(1) \textbf{One-step retrieval}: By keeping all other components of our pipeline unchanged, we reduce the current two-step retrieval method to a one-step pipeline by directly retrieving the top-12 chunks from the entire paper and GitHub documents without first inferring relevant paragraphs. Although intuitively the nature of our question template set correlates closely with the sectional structure of papers and GitHub repositories, this baseline could provide further support of using a paragraph-level retriever.

(2) \textbf{Retrieval only}: Upon completing the two-step retrieval and obtaining relevant chunks, the method directly use the retrieved chunk as the final output. This is used to assess the advantages of the summary generation step over merely using the author's original text.

We compare our \ourmethod pipeline with these two baselines in \cref{sec:baseline_result}.

\section{Evaluation Setup}

\begin{table*}[t!]
    \centering
    \resizebox{\textwidth}{!}{
    \begin{tabular}{lllllllll}
    \toprule
    \textbf{Metric} & \textbf{Human} & \textbf{Claude3 Opus} & \textbf{GPT4} & \textbf{GPT3.5} & \textbf{Llama2 70B} & \textbf{Vicuna 13B} & \textbf{Llama2 7B} & \textbf{Mistral 7B} \\
    \hline
    Completeness & 1.92 & \textbf{7.28} & 5.99 & 5.24 & 4.76 & 4.24 & 2.50 & 4.07 \\
    Accuracy & \textbf{6.66} & 6.56 & 6.04 & 4.51 & 3.61 & 3.11 & 1.84 & 3.67 \\
    Objectivity & 2.03 & \textbf{7.16} & 6.33 & 5.23 & 4.72 & 4.25 & 2.12 & 4.16 \\
    Understandability & 2.49 & \textbf{7.11} & 6.21 & 4.99 & 4.80 & 3.80 & 2.09 & 4.51 \\
    Reference quality & 6.13 & \textbf{6.75} & 5.40 & 4.28 & 4.15 & 3.73 & 1.63 & 3.93 \\
    \bottomrule
    \end{tabular}
    }
    \caption{Human evaluation results on LLM generated and human-generated model cards.}
    \label{tab:human_eval}
\end{table*}

We evaluate \ourmethod on various standard as well as state-of-the-art metrics to measure the faithfulness, relevance, and other aspects of the generation quality. Additionally, we incorporate human evaluation for the pipeline to address three key challenges that can't be solved by automatic metrics alone: First, there is an absence of ground truth labels of generated model cards by \ourmethod. To mitigate this, we have to develop specific manual evaluations to assess performance. Second, current model cards created by human developers are often incomplete and  deviate from the recommended template provided by HF. Third, the LLM generated model card is typically long with over 4000 words, and brings challenges to both open-source standard evaluations with limited context size and costly GPT-based metrics. 

\paragraph{Standard Metrics}
We follow \citet{honovich2022true} and use ROUGE \citep{lin-2004-rouge}, BERTScore \citep{zhang2019bertscore}, BARTScore \citep{yuan2021bartscore}, and NLI-finetuned models \citep{williams-etal-2018-broad, maccartney-manning-2008-modeling} to measure the factual consistency of retrieved chunks set $\bm{R}$ and the generated answer $\bm{A}$. Due to the large size of retrieved texts, we use \texttt{deberta-v3-base} as the base model for BERTScore, and use \texttt{nli-deberta-v3-large} as the NLI-finetuned model scorer \citep{reimers2019sentence, he2021debertav3}. More details in \cref{appn:metrics}.

\paragraph{GPT Metrics}
Following \citet{es2023ragas}, we consider the measurement of faithfulness, answer relevance, context precision, and context relevance using GPT4. \cref{tab:gpt_metric} provides a description of these metrics. As different combinations of inputs are taken into consideration, these metrics are necessary supplements to standard metrics. Full prompt details are explained in \cref{appn:metrics}.

\paragraph{Human Evaluation Metrics}
Putting together LLM generated cards with the human-generated cards as a sample, we devise the following manual evaluation metrics: completeness, accuracy, objectivity, understandability, and reference quality. We design a simple Gradio annotation interface 
\citep{abid2019gradio}, and more details are in \cref{appn:human_annotate}.

\section{Results}

\subsection{Performance Summary}
Our human evaluation results are shown in \cref{tab:human_eval} and automatic evaluation results are shown in \cref{tab:eval_hallucination,tab:eval_gpt4} for model cards. The only difference for the data card generation pipeline is the substitution with data card question templates. 
In this subsection, we mainly answer two questions below:

\begin{table*}[t]
    \centering
    \resizebox{\textwidth}{!}{
    \begin{tabular}{l|l|ccccccc|c}
    \toprule
    \textbf{Metric} & \textbf{Model} & \textbf{Summary} & \textbf{Model details} & \textbf{Uses} & \textbf{Bias} & \textbf{Training details} & \textbf{Evaluation} & \textbf{More info} & \textbf{All} \\
    \hline
    \multirow{7}{*}{\textbf{ROUGE-L}} & Claude3 Opus & 8.91 & 11.26 & 14.39 & 14.21 & 14.11 & 14.99 & 12.78 & 13.04 \\
     & GPT4 & 8.80 & 9.35 & 15.38 & 18.20 & 17.59 & \textbf{19.40} & 9.73 & 13.27 \\
     & GPT3.5 & 9.90 & 10.70 & \textbf{16.51} & \textbf{20.21} & 14.46 & 15.75 & 10.73 & 13.16 \\
     & Llama2 70b chat & \textbf{12.71} & \textbf{14.35} & 12.85 & 17.20 & \textbf{18.74} & 18.03 & \textbf{16.21} & \textbf{15.98} \\
     & Vicuna 13b v1.5 & 10.78 & 11.35 & 13.54 & 17.10 & 16.06 & 16.75 & 10.29 & 13.12 \\
     & Llama2 7b chat & 11.91 & 12.84 & 13.89 & 15.85 & 14.63 & 16.21 & 13.61 & 14.08 \\
     & Mistral 7b inst & 12.19 & 11.01 & 13.02 & 15.07 & 16.79 & 16.23 & 9.47 & 12.70 \\
    \hline
    \multirow{7}{*}{\textbf{BERTScore}} & Claude3 Opus & 54.78 & 53.73 & 58.42 & 56.32 & 57.83 & 58.80 & 55.17 & 56.10 \\
     & GPT4 & 54.06 & 50.44 & 57.81 & 58.81 & 59.50 & \textbf{61.24} & 47.48 & 53.96 \\
     & GPT3.5 & 54.86 & 53.17 & \textbf{58.62} & \textbf{59.29} & 56.61 & 57.42 & 52.47 & 55.09 \\
     & Llama2 70b chat & \textbf{57.21} & \textbf{56.15} & 53.97 & 56.55 & \textbf{59.69} & 59.46 & \textbf{56.99} & \textbf{57.21} \\
     & Vicuna 13b v1.5 & 55.15 & 52.97 & 54.99 & 57.24 & 57.61 & 58.83 & 52.10 & 54.83 \\
     & Llama2 7b chat & 55.76 & 54.51 & 53.93 & 55.48 & 56.30 & 57.13 & 54.72 & 55.26 \\
     & Mistral 7b inst & 55.69 & 52.80 & 54.12 & 53.76 & 57.10 & 57.63 & 49.12 & 53.47 \\
    \hline
    \multirow{7}{*}{\textbf{BARTScore}} & Claude3 Opus & 13.92 & 5.60 & \textbf{2.56} & 1.59 & 4.10 & 2.87 & 4.33 & 4.36 \\
     & GPT4 & 9.69 & 7.63 & 1.43 & 1.98 & 4.02 & 4.29 & 6.11 & 5.34 \\
     & GPT3.5 & \textbf{17.09} & 9.58 & 2.04 & 3.52 & 5.75 & 6.65 & \textbf{9.10} & 7.61 \\
     & Llama2 70b chat & 14.17 & 5.41 & 1.45 & 3.10 & 5.30 & 4.60 & 5.91 & 5.15 \\
     & Vicuna 13b v1.5 & 13.53 & 5.67 & 1.90 & \textbf{3.76} & 5.63 & 6.81 & 6.77 & 5.90 \\
     & Llama2 7b chat & 14.04 & 3.49 & 2.11 & 3.61 & 4.70 & 3.68 & 4.01 & 4.03 \\
     & Mistral 7b inst & 16.52 & \textbf{9.65} & 2.00 & 3.55 & \textbf{7.00} & \textbf{8.75} & 8.31 & \textbf{7.90} \\
    \hline
    \multirow{7}{*}{\textbf{NLI}} & Claude3 Opus & 58.00 & \textbf{54.62} & 56.33 & 59.00 & 62.25 & 61.40 & 60.12 & 58.68 \\
     & GPT4 & 61.00 & 52.88 & 53.00 & 56.00 & \textbf{64.50} & \textbf{65.60} & \textbf{62.62} & \textbf{59.42} \\
     & GPT3.5 & \textbf{65.14} & 49.83 & 57.54 & \textbf{62.41} & 59.14 & 60.14 & 56.80 & 56.54 \\
     & Llama2 70b chat & 56.46 & 51.70 & 55.22 & 58.42 & 57.70 & 62.04 & 59.74 & 57.14 \\
     & Vicuna 13b v1.5 & 60.20 & 51.40 & \textbf{58.05} & 55.10 & 58.29 & 63.33 & 55.00 & 56.31 \\
     & Llama2 7b chat & 56.46 & 50.19 & 54.31 & 57.23 & 57.82 & 62.11 & 56.44 & 55.77 \\
     & Mistral 7b inst & 58.67 & 50.36 & 54.25 & 54.59 & 59.06 & 58.91 & 55.17 & 55.02 \\
    \bottomrule
    \end{tabular}
    }
    \caption{Factual consistency evaluation results per section on our retrieve-and-generate pipeline using ROUGE-L, BERTScore, BARTScore, and NLI pretrained scorers.}
    \label{tab:eval_hallucination}
\end{table*}

\begin{table*}[t]
    \centering
    \resizebox{\textwidth}{!}{
    \begin{tabular}{l|l|ccccc}
    \toprule
    \textbf{Metric} & \textbf{Model} & \textbf{Summary} & \textbf{Description} & \textbf{Direct use} & \textbf{Bias, risks, limitation} & \textbf{Results summary} \\
    \hline
    \multirow{7}{*}{\textbf{Faithfulness}} & Claude3 Opus & 74.97 & 49.77 & \textbf{78.23} & \textbf{71.28} & 84.89 \\
     & \textbf{GPT4} & 68.87 & \textbf{85.58} & 62.99 & 64.20 & \textbf{86.44} \\
     & GPT3.5 & 71.23 & 83.21 & 48.71 & 55.17 & 82.99 \\
     & Llama2 70b chat & 70.03 & 76.39 & 43.20 & 32.14 & 63.87 \\
     & Vicuna 13b v1.5 & \textbf{78.46} & 81.74 & 45.94 & 46.64 & 78.22 \\
     & Llama2 7b chat & 72.41 & 71.35 & 48.43 & 44.23 & 65.56 \\ 
     & Mistral 7b inst & 76.75 & 75.03 & 38.28 & 41.77 & 73.61 \\
    \hline
    \multirow{7}{*}{\textbf{Answer relevance}} & Claude3 Opus & 90.42 & 91.10 & 89.12 & 91.39 & 93.15 \\
     & GPT4 & 90.83 & 93.12 & 89.69 & 92.03 & 91.36 \\
     & \textbf{GPT3.5} & \textbf{91.18} & \textbf{93.26} & 90.70 & \textbf{93.75} & \textbf{93.24} \\
     & Llama2 70b chat & 90.76 & 92.27 & 91.25 & 92.23 & 91.63 \\
     & Vicuna 13b v1.5 & 89.00 & 91.22 & 90.17 & 92.99 & 90.38 \\
     & Llama2 7b chat & 90.44 & 90.95 & \textbf{92.55} & 92.69 & 92.81 \\
     & Mistral 7b inst & 90.46 & 91.77 & 90.36 & 91.56 & 90.43 \\
    \hline
     \multirow{7}{*}{\textbf{Context precision}} & Claude3 Opus & 33.25 & 51.73 & 26.17 & 20.99 & 42.58 \\
     & \textbf{GPT4} & \textbf{35.01} & 51.25 & \textbf{29.29} & \textbf{22.76} & 40.23 \\
     & GPT3.5 & 29.07 & 51.80 & 25.71 & 18.77 & 37.88 \\
     & Llama2 70b chat & 21.05 & 50.00 & 25.35 & 20.03 & 40.82 \\
     & Vicuna 13b v1.5 & 24.91 & 51.22 & 24.00 & 8.93 & 39.00 \\
     & Llama2 7b chat & 32.46 & 50.79 & 25.52 & 14.27 & 40.04 \\
     & Mistral 7b inst & 31.10 & \textbf{52.22} & 28.45 & 21.36 & \textbf{44.45} \\
    \hline
    \multirow{7}{*}{\textbf{Context relevance}} & Claude3 Opus & 13.32 & 48.82 & 28.90 & \textbf{21.32} & 23.01 \\
     & GPT4 & 12.86 & \textbf{52.39} & 26.63 & 18.89 & 23.47 \\
     & \textbf{GPT3.5} & 13.27 & 51.03 & \textbf{29.82} & 18.97 & \textbf{26.44} \\
     & Llama2 70b chat & 13.32 & 49.62 & 27.22 & 18.37 & 24.31 \\
     & Vicuna 13b v1.5 & 13.83 & 51.32 & 27.00 & 14.03 & 23.08 \\
     & Llama2 7b chat & \textbf{13.87} & 50.78 & 28.07 & 17.57 & 26.23 \\
     & Mistral 7b inst & 13.22 & 47.05 & 28.40 & 18.75 & 23.52 \\
    \bottomrule
    \end{tabular}
    }
    \caption{GPT4 evaluation results on five most important questions based on faithfulness (Faith), answer relevance (AR), context precision (CP), and context relevance (CR).}
    \label{tab:eval_gpt4}
\end{table*}

\paragraph{Are our generated model cards better than human-generated ones?}
We conduct a random sampling of 50 model cards from the test set and compute the average metric scores across all the annotated samples, as shown in \cref{tab:human_eval}. GPT3.5, GPT4, and Claude3 Opus demonstrates superior performance over other open-sourced LLMs and human-generated content in terms of completeness, objectivity, and understandability. This finding aligns with the observations presented below for \cref{tab:eval_hallucination,tab:eval_gpt4}.

Conversely, the human-generated model cards often received higher scores in accuracy and reference quality. This disparity suggests that all LLMs exhibit some degree of hallucination for factual content and reference links in their generation. It is important to note that the human-generated model cards' incompleteness precludes a direct comparison of human evaluation metrics with the metrics used in \cref{tab:eval_hallucination,tab:eval_gpt4}. Moreover, the insights derived from \cref{tab:human_eval} are not obtainable through automatic metrics. We thus conclude that human evaluation metrics are indispensable components of our overall evaluation framework.

\paragraph{How does GPT3.5 perform compared with open sourced LLMs?}
From \cref{tab:eval_hallucination}, we can't observe a uniform trend for factual consistency across all sub-tasks. GPT3.5 outperforms open-sourced LLMs on ``Uses'' and ``Bias'' question sets in 3 over 4 
standard metrics, while Llama2 70b generates more factual consistent answers on other sub-tasks according to ROUGE-L and BERTScore.

According to \cref{tab:eval_gpt4}, GPT3.5 beats other LLMs on faithfulness and answer relevance across nearly all sub-tasks, and shows its strong instruction-following capabilities for question-answering. However, we have an interesting observation that though GPT3.5 has higher context relevance scores, it is outperformed by Mistral 7B on context precision. A higher context relevance indicates that the question can be better answered from the given context, while a lower context precision means that the context may contain other unnecessary information for answering the question. The discrepancy between results by these two metrics suggests that retrieved texts from the GPT \ourmethod pipeline are more informative but less concise. Additionally, since we use LLM generated pseudo answers as queries for similar paragraphs, pseudo answers with more possibly unrelated contents will lead to more irrelevant chunks from retrieval. Along with the illustration in Appendix \cref{fig:pseudo_answer_len}, we draw the conclusion that GPT3.5 generates pseudo answers with potentially more unrelated details.

\subsection{Baseline Results}
\label{sec:baseline_result}

To assess the effectiveness of \ourmethod's retriever and generator, we first compare it to the baseline methods outlined in \cref{sec:baseline}. To manage the expenses associated with OpenAI AI calling, we employ GPT3.5 for subsequent studies. We obtain Krippendorff's $\alpha$ (mean=0.83, std=0.14, min=0.56, max=0.99) for the agreements on \cref{tab:eval_gpt4} by GPT4 and GPT3.5 to validate our evaluation model substitution \citep{castro-2017-fast-krippendorff}. 

\begin{table}[ht]
    \centering
    \resizebox{\columnwidth}{!}{
    \begin{tabular}{l|l|ll}
    \toprule
    \textbf{Model} & \textbf{Method} & \textbf{CP} & \textbf{CR} \\
    \hline
    \multirow{2}{*}{GPT3.5} & One-step retrieval & 44.03 & 27.82 \\
    & \ourmethod & 44.67 (\green{+0.64}) & 28.24 (\green{+0.42}) \\
    \hline
    \multirow{2}{*}{Llama2 70B} & One-step retrieval & 42.94 & 28.10 \\
    & \ourmethod & 44.03 (\green{+1.09}) & 28.83 (\green{+0.73}) \\
    \hline
    \multirow{2}{*}{Llama2 7B} & One-step retrieval & 43.47 & 27.35 \\
    & \ourmethod & 41.91 (\red{-1.56}) & 28.00 (\green{+0.65}) \\
    \hline
    \multirow{2}{*}{Mistral 7B} & One-step retrieval & 43.75 & 27.80 \\
    & \ourmethod & 45.24 (\green{+1.49}) & 27.97 (\green{+0.17}) \\
    \bottomrule
    \end{tabular}
    }
    \caption{GPT3.5 evaluation results of the one-step retrieval baseline and \ourmethod in terms of context precision and context relevance.}
    \label{tab:baseline1}
\end{table}

\vspace{-0.5em}
\paragraph{One-step retrieval} Since the change is only in the retrieval process in comparison to \ourmethod, we focus exclusively on context precision and context relevance as metrics. These metrics evaluate the quality of the retrieved text $r_i$ in response to a given question $q_i$. We evaluate across four LLMs, and report results based on the averaged score of the most important questions. According to \cref{tab:baseline1}, the two-step retrieval process achieves marginally yet consistently higher scores than the one-step retrieval across nearly all models. These findings indicate that a paragraph-level retrieval model constitutes a more appropriate method for this study.

\begin{table}[ht]
    \centering
    \resizebox{\columnwidth}{!}{
    \begin{tabular}{l|l|lc}
    \toprule
    \textbf{Model} & \textbf{Method} & \textbf{AR} & \textbf{Understandability} \\
    \hline
    \multirow{2}{*}{GPT3.5} & Retrieval only & 81.28 & 5.60\% \\
    & \ourmethod & 90.84 (\green{+9.56}) & \textbf{94.40\%} \\
    \hline
    \multirow{2}{*}{Llama2 70B} & Retrieval only & 81.61 & 1.60\% \\
    & \ourmethod & 90.32 (\green{+8.71}) & \textbf{98.40\%} \\
    \hline
    \multirow{2}{*}{Llama2 7B} & Retrieval only & 81.32 & 4.40\% \\
    & \ourmethod & 90.78 (\green{+9.46}) & \textbf{95.60\%} \\
    \hline
    \multirow{2}{*}{Mistral 7B} & Retrieval only & 81.49 & 2.40\% \\
    & \ourmethod & 89.83 (\green{+8.34}) & \textbf{97.60\%} \\
    \bottomrule
    \end{tabular}
    }
    \caption{GPT3.5 evaluation results of the retrieval-only baseline and \ourmethod in terms of answer relevance and understandability. Full results including assessments of brevity can be found in Appendix \cref{tab:baseline2_full}.}
    \label{tab:baseline2}
\end{table}

\begin{table*}[t!]
    \centering
    \resizebox{0.95\textwidth}{!}{
    \begin{tabular}{l|l|lllll}
    \toprule
    \textbf{Metric}& \textbf{Model} & \textbf{Summary} & \textbf{Description} & \textbf{Direct use} & \textbf{Bias, risks, limitation} & \textbf{Results summary} \\
    \hline
    \multirow{2}{*}{\textbf{NLI}} & GPT3.5 & 65.14 (\green{+2.14}) & 51.53 (\green{+0.53}) & 50.51 (\green{+0.51}) & 64.12 (\green{+1.12}) & 58.50 (\green{+0.50}) \\
    & w/o pseudo & 63.00 & 51.00 & 50.00 & 63.00 & 58.00 \\
    \hline
    \multirow{2}{*}{\textbf{Faith}}
    & GPT3.5 & 81.93 (\green{+6.75}) & 79.30 (\green{+4.30}) & 41.23 (\green{+0.62}) & 46.42 (\rred{-2.53}) & 72.66 (\green{+1.21}) \\
     & w/o pseudo & 75.18 & 75.00 & 40.61 & 48.95 & 71.45 \\
    \hline
    \multirow{2}{*}{\textbf{AR}}
    & GPT3.5 & 86.94 (\green{+0.06}) & 89.56 (\rred{-0.65}) & 88.95 (\green{+0.78}) & 93.55 (\green{+0.40}) & 95.20 (\green{+0.02}) \\
     & w/o pseudo & 86.88 & 90.21 & 88.17 & 93.15 & 95.18 \\
    \hline
    \multirow{2}{*}{\textbf{CP}}
    & GPT3.5 & 47.53 (\green{+7.49}) & 19.61 (\green{+1.01}) & 13.44 (\green{+3.20}) & 13.03 (\rred{-0.26}) & 64.15 (\green{+0.24}) \\
     & w/o pseudo & 40.04 & 18.60 & 10.24 & 13.29 & 63.91 \\
    \hline
    \multirow{2}{*}{\textbf{CR}}
    & GPT3.5 & 11.85 (\green{+2.32}) & 23.24 (\rred{-2.21}) & 8.70 (\green{+1.19}) & 4.35 (\green{+0.69}) & 24.04 (\green{+5.79}) \\
     & w/o pseudo & 9.53 & 25.45 & 7.51 & 3.66 & 18.25 \\
     \hline\hline
        \multirow{2}{*}{\textbf{Faith}} & GPT3.5 & 81.93 (\green{+8.09}) & 79.30 (\green{+15.31}) & 41.23 (\green{+26.62}) & 46.42 (\green{+22.14}) & 72.66 (\green{+25.16}) \\
         & Llama2 70B & 73.84 & 63.99 & 14.61 & 24.28 & 47.50 \\
        \hline
        \multirow{2}{*}{\textbf{AR}} & GPT3.5 & 86.94 (\rred{-1.56}) & 89.56 (\green{+0.63}) & 88.95 (\green{+6.58}) & 93.55 (\green{+9.53}) & 95.20 (\green{+7.21}) \\
         & Llama2 70B & 88.50 & 88.93 & 82.37 & 84.02 & 87.99 \\
    \bottomrule
    \end{tabular}
    }
    \caption{GPT3.5 evaluation results on five most important questions for pseudo answer chain ablation in top five rows and generation chain ablation in bottom two rows. For the generation chain ablation, we keep all previous chains unchanged with \texttt{GPT-3.5-turbo} as the backbone, and only vary the choice of LLMs for the final generation chain, including \texttt{GPT-3.5-turbo} and \texttt{Llama2-70B-Chat-HF}.}
    \label{tab:ablation_pseudo_answer}
\end{table*}

\vspace{-0.5em}
\paragraph{Retrieval only} Following the same evaluation setup as above, we consider answer relevance of generated text $g_i$ according to a given question $q_i$. To further compare which method produces more understandable and concise outputs, we also incorporate understandability and brevity into our evaluation as GPT metrics for pairwise comparison \citep{liusie-etal-2024-llm, fu2023gptscore}. As illustrated in \cref{tab:baseline2}, \ourmethod significantly outperforms the retrieval-only baseline across all metrics, highlighting the importance of the generation step in summarizing and restating sentences from source documents to enhance their understandability and conciseness. Further details are in \cref{appn:baseline2}. 


\subsection{Ablation Study}
We also conducted the following ablation studies and explored model architecture variations to further validate \ourmethod's components: (1) Remove the pseudo answer chain and use original questions for embedding similarity matching. (2) Vary the final generation chain only with different LLMs, and maintain all preceding reasoning chains as generated by GPT3.5. (3) Employ different embedding models for dense retrieval.

\paragraph{Pseudo Answer Chain}
We compare the GPT evaluation scores and factual consistency using NLI of \ourmethod + GPT3.5 pipeline with or without the pseudo answer chain, as illustrated in \cref{tab:ablation_pseudo_answer}. \ourmethod with the pseudo answer chain outperforms the other across nearly all important questions and metrics being tested. Our results demonstrate the necessity  of the pseudo answer chain in our pipeline. Some lower scores may be because of more unrelated texts from the generated pseudo answers for specific questions.

\paragraph{Generation Chain}
In bottom two rows of \cref{tab:ablation_pseudo_answer}, we show the comparison results by only substituting GPT3.5 in the generation chain with Llama2 70B based on faithfulness and answer relevance. Context precision and context relevance are the same since retrieved texts remain unchanged. We observe a large drop for the faithfulness score and a moderate drop for the answer relevance score, indicating the stronger instruction following capability of GPT3.5 in the generation stage compared to Llama2 70B.


\paragraph{Embedding Models}
\label{par:embed}
We compare the embedding model \texttt{jina-embeddings-v2-base-en} that we use with two other commonly used sentence transformer models: \texttt{all-MiniLM-L6-v2} and \texttt{all-mpnet-base-v2} \citep{günther2023jina,wang2020minilm,reimers-2019-sentence-bert,reimers-2020-multilingual-sentence-bert}. We justify our choice of embedding models in Appendix \cref{fig:ablation_embed}, where \ourmethod with \texttt{jina-embeddings-v2-base-en} performs better than others according to all three metrics related to the retrieved texts.


\subsection{LLM Generated Model Card Statistics}
\cref{appn:gen_stats} provides related statistics. Compared with statistics in \cref{tab:stat_card}, LLM generates longer and more informative than human.

\section{Conclusion}
In this study, we introduce a novel task focused on the automatic generation of model cards and data cards. This task is facilitated by the creation of the \ourdata dataset, and the development of the \ourmethod pipeline  leveraging state-of-the-art LLMs. The system is designed to assist in the generation of understandable, comprehensive, and consistent models and data cards, thereby providing a valuable contribution to the field of responsible AI.

\section*{Limitations}

One limitation of our method is that, despite the adoption of the RAG pipeline and explicit instructions for LLMs to adhere closely to the retrieved text, there remains the potential for hallucinations in the generated text. To mitigate this, future work may integrate specific strategies into our \ourmethod pipeline for hallucination reduction by carefully balancing generation speed with quality.

Our current approach employs a single-step generation process and a two-step retrieval process that first infers relevant section contents. Future work could incorporate more advanced chain-of-thought prompting techniques and compare with our \ourmethod pipeline. For complex questions requiring multistep reasoning, after decomposed into manageable sub-questions, we can address each sub-question through multiple reasoning steps, as suggested by recent research \citep{yao2022react,khot2022decomposed,press2022measuring,he2022rethinking}.
Additionally, an iterative retrieval-generation collaborative framework can also be used to refine responses in each iteration based on newly retrieved contexts, following recent advancements in iterative retrieval and generation frameworks for complex tasks \citep{shao2023enhancing,feng2023retrieval}.

\section*{Ethical Considerations}
%

This work aims to provide insights about the automatic generation of model cards and data cards. Such an endeavor is instrumental in promoting accountability and traceability among developers as they document their models. The dataset for this research was collected using public REST APIs from HF Hub, Arxiv, and GitHub. We ensured that only open-source model cards, data cards, and their associated source documents were collected, strictly adhering to the stipulations of their respective licenses for research purposes, so there were no user privacy concerns in the dataset. Our dataset and method should only be used for research purpose.

On the other hand, while the questions we pose to LLMs are technical and specific, there remains a risk of receiving biased responses, particularly for certain queries. For instance, the question about model limitations might yield biased answers, as source papers and GitHub READMEs could contain overstated claims about their models. Consequently, our generated model cards could contain these statements as well if the source texts containing them are retrieved.

To mitigate this, one reasonable approach is to insert a step after retrieval to filter out or neutralize overstatements. Additionally, we can explicitly prompt LLMs to account for such biases during the generation stage. Another concern is the potential for content homogeneity when using LLMs for model card generation. Excessive reliance on templates could limit model card creators’ potential to discuss new issues not covered in the original papers or GitHub repositories \citep{PMID:37349357,PMID:37985919}.

Moreover, one aspect of our approach is that we use direct prompts to LLMs rather than fine-tuning them on human-generated model cards, which can also exhibit biases from the internal of LLMs, such as overstatements on well-known models or omissions of potential risks. In our analysis of 2495 human-written model cards in our dataset, only 30.54\% mention “weakness(es)” or “limitation(s)”, and 15.23\% mention “bias(es)”. If future study can collect more fairly-written human-generated model cards, they can also be used to finetune LLMs for better performance on this task.

\bibliography{main}
\clearpage
\appendix

\section{Question Templates}
\label{appn:question_template}

\begin{table*}[t!]
    \centering \small
    \begin{tabularx}{\textwidth}{llX}
    \toprule
    \textbf{Question} & \textbf{Role} & \textbf{Prompt} \\
\hline
Summary & Project organizer & Provide a 1-2 sentence summary of what the model is. \\
Description & Project organizer & Provide basic details about the model. This includes the model architecture, training procedures, parameters, and important disclaimers. \\
Funded by & Project organizer & List the people or organizations that fund this project of the model. \\
Shared by & Developer & Who are the contributors that made the model available online as a GitHub repo? \\
Model type & Project organizer & Summarize the type of the model in terms of the training method, machine learning type, and modality in one sentence. \\
Language & Project organizer & Summarize what natural human language the model uses or processes in one sentence. \\
License & Project organizer & Provide the name and link to the license being used for the model. \\
Finetuned from & Project organizer & If the model is fine-tuned from another model, provide the name and link to that base model. \\
Demo sources & Project organizer & Provide the link to the demo of the model. \\
Direct use & Project organizer & Explain how the model can be used without fine-tuning, post-processing, or plugging into a pipeline. Provide a code snippet if necessary \\
Downstream use & Project organizer & Explain how this model can be used when fine-tuned for a task or when plugged into a larger ecosystem or app. Provide a code snippet if necessary \\
Out of scope use & Sociotechnic & How the model may foreseeably be misused and address what users ought not do with the model. \\
Bias risks limitations & Sociotechnic & What are the known or foreseeable issues stemming from this model? These include foreseeable harms, misunderstandings, and technical and sociotechnical limitations. \\
Bias recommendations & Sociotechnic & What are recommendations with respect to the foreseeable issues about the model? \\
Training data & Developer & Write 1-2 sentences on what the training data of the model is. Links to documentation related to data pre-processing or additional filtering may go here as well as in More Information. \\
Preprocessing & Developer & Provide detail tokenization, resizing/rewriting (depending on the modality), etc. about the preprocessing for the data of the model. \\
Training regime & Developer & Provide detail training hyperparameters when training the model. \\
Speeds sizes times & Developer & Provide detail throughput, start or end time, checkpoint sizes, etc. about the model. \\
Testing data & Developer & Provide benchmarks or datasets that the model evaluates on. \\
Testing factors & Sociotechnic & What are the foreseeable characteristics that will influence how the model behaves? This includes domain and context, as well as population subgroups. Evaluation should ideally be disaggregated across factors in order to uncover disparities in performance. \\
Testing metrics & Developer & What metrics will be used for evaluation in light of tradeoffs between different errors about the model? \\
Results & Developer & Provide evaluation results of the model based on the Factors and Metrics. \\
Results summary & Developer & Summarize the evaluation results about the model. \\
Model examination & Developer & This is an experimental section some developers are beginning to add, where work on explainability/interpretability may go about the model. \\
Hardware & Developer & Provide the hardware type that the model is trained on. \\
Software & Developer & Provide the software type that the model is trained on. \\
Hours used & Developer & Provide the amount of time used to train the model. \\
Cloud provider & Developer & Provide the cloud provider that the model is trained on. \\
Co2 emitted & Developer & Provide the amount of carbon emitted when training the model. \\
Model specs & Developer & Provide the model architecture and objective about the model. \\
Compute infrastructure & Developer & Provide the compute infrastructure about the model. \\
    \bottomrule
    \end{tabularx}
    \caption{Template of the all questions necessary for generating a whole model card.}
    \label{tab:qt_model_card}
\end{table*}

\begin{table*}[t!]
    \centering \small
    \begin{tabularx}{\textwidth}{llX}
\toprule
Question & Role & Prompt \\
\hline
Description & Data manager & Provide the homepage link for the dataset, just give me a link please. \\
Leaderboard & Data manager & Provide the Leaderboard link for the dataset. \\
Pointofcontact & Data manager & Provide the Point of Contact for the dataset. \\
Summary & Data manager & Provide basic details about the dataset. Briefly summarize the dataset, its intended use and the supported tasks. Give an overview of how and why the dataset was created. The summary should explicitly describe the domain, topic, or genre covered. \\
Supported tasks and leaderboards & Data analyst & Describe the tasks and leaderboards supported by the dataset. Include task description, metrics, suggested models, and leaderboard details. \\
Languages & Data analyst & Provide an overview of the languages represented in the dataset, including details like language type, script, and region. Include BCP-47 codes if available. \\
Data instances & Data scientist & Provide a JSON-formatted example of a typical instance in the dataset with a brief description. Include a link to more examples if available. Describe any relationships between data points. \\
Data fields & Data architect & List and describe the fields in the dataset, including their data type, usage in tasks, and attributes like span indices. Mention if the dataset contains example IDs and their inherent meaning. \\
Data splits & Data manager & Describe the data splits in the dataset. Include details such as the number of splits, any criteria used for splitting the data, differences between the splits, and the sizes of each split. Provide descriptive statistics for the features where appropriate, for example, average sentence length for each split. \\
Curation rationale & Data manager & What need or purpose motivated the creation of this dataset? Describe the underlying reasons and major choices involved in its assembly. Explain the significance of the dataset in its field and any specific gaps or demands it aims to address. \\
Source data & Data manager & Describe the source data used for this dataset. Describe the data collection process. Describe any criteria for data selection or filtering. List any key words or search terms used. If possible, include runtime information for the collection process. \\
Source language producers & Data manager & Clarifying the human or machine origin of the dataset. Avoiding assumptions about the identity or demographics of the data creators. Providing information about the people represented in the data, with references where applicable. \\
Annotations & Data manager & Describe the annotation process to the dataset.   Detail the annotation process and tools used, or note if none were applied.   Specify the volume of data annotated.  \\
Annotators & Data manager & Describe the annotator of the dataset. For annotations in the dataset, state their human or machine-generated nature.  Describe the creators of the annotations, their selection process, and any self-reported demographic information. \\
Personal and sensitive information & Data manager & Categorize how identity data, such as gender referencing Larson (2017), is sourced and used in the dataset. Indicate if the data includes sensitive information or can identify individuals. Describe any anonymization methods applied.\\
Social impact of dataset & Data manager & Explore the dataset's social impacts: its role in advancing technology and enhancing quality of life. Consider negative effects like decision-making opacity and reinforcing biases. Check if it includes low-resource or under-represented languages. Assess its impact on underserved communities.\\
Discussion of biases & Data manager & When constructing datasets, especially those including text-based content like Wikipedia articles, biases may be present. If there have been analyses to quantify these biases, it's important to summarize these studies and note any measures taken to mitigate the biases. \\
Other known limitation & Data analyst & Outline and cite any known limitations of the dataset, such as annotation artifacts, in your studies. \\
Dataset curators & Data manager & List the people involved in collecting the dataset and their affiliations. If known, include information about funding sources for the dataset. This should encompass individuals, organizations, and any collaborative efforts involved in the dataset creation. \\
Licensing information & Legal advisor & Provide the license and link to the license webpage if available for the dataset. \\
Contributions & Data manager & Write in 1-2 sentence about the contributers for the dataset. Mention the GitHub username and provide their GitHub profile link. You should follows the format: Thanks to [@github-username](https://github.com/<github-username>) for adding this dataset. \\
\bottomrule
\end{tabularx}
    \caption{Template of the all questions necessary for generating a whole data card.}
    \label{tab:qt_dataset_card}
\end{table*}

\cref{tab:qt_model_card,tab:qt_dataset_card} shows full question templates of model cards and data cards. We have 31 questions in total for generating model cards, and 21 questions for generating data cards. We create these questions based on the template provided by HF,\footnote{\url{https://github.com/huggingface/huggingface_hub/tree/main/src/huggingface_hub/templates}} and include necessary requirements. 

\section{Dataset Collection Details}
\label{appn:data_collect}

\begin{figure}[h]
    \centering
    \includegraphics[width=\columnwidth]{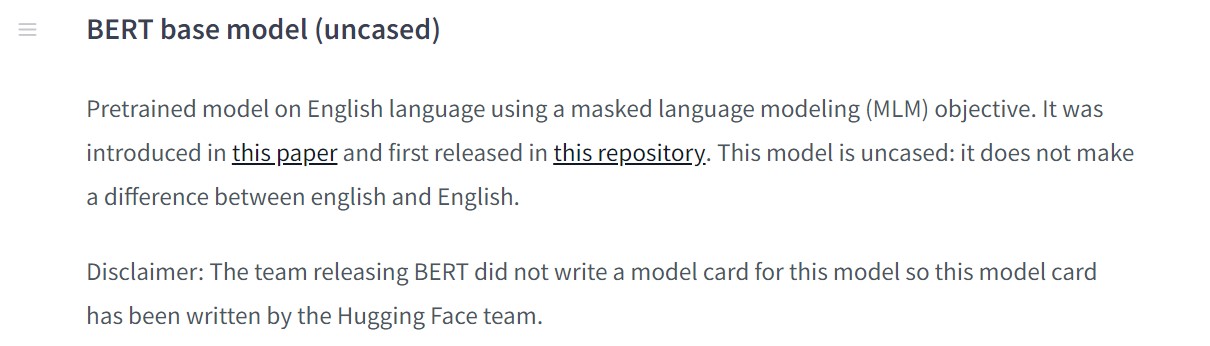}
    \caption{\texttt{bert-base-uncased} \citep{devlin2018bert} as a current model card example with a unique disclaimer sentence, indicating a modification by the HF team.}
    \label{fig:screenshot_bert_disclaimer}
\end{figure}

For the model card evaluation set selection, we select all 350 examples that are rewritten by the HF team with their unique disclaimers, as shown in \cref{fig:screenshot_bert_disclaimer}.

\section{Dataset Annotation Details}
\label{appn:data_annotate}

\paragraph{Human Annotation Guidelines} To evaluate paper links and direct GitHub links on the model card evaluation set, we require the annotators to go through each current model card and provide all possible paper links and GitHub links to annotators. They are asked to select the direct paper link and GitHub link from all candidate links, by looking at their positions of occurrences in the model card example. If no direct links of either sources can be determined, they need to label this model card as ``Invalid''.

\begin{figure}[H]
    \centering
    \includegraphics[width=\columnwidth]{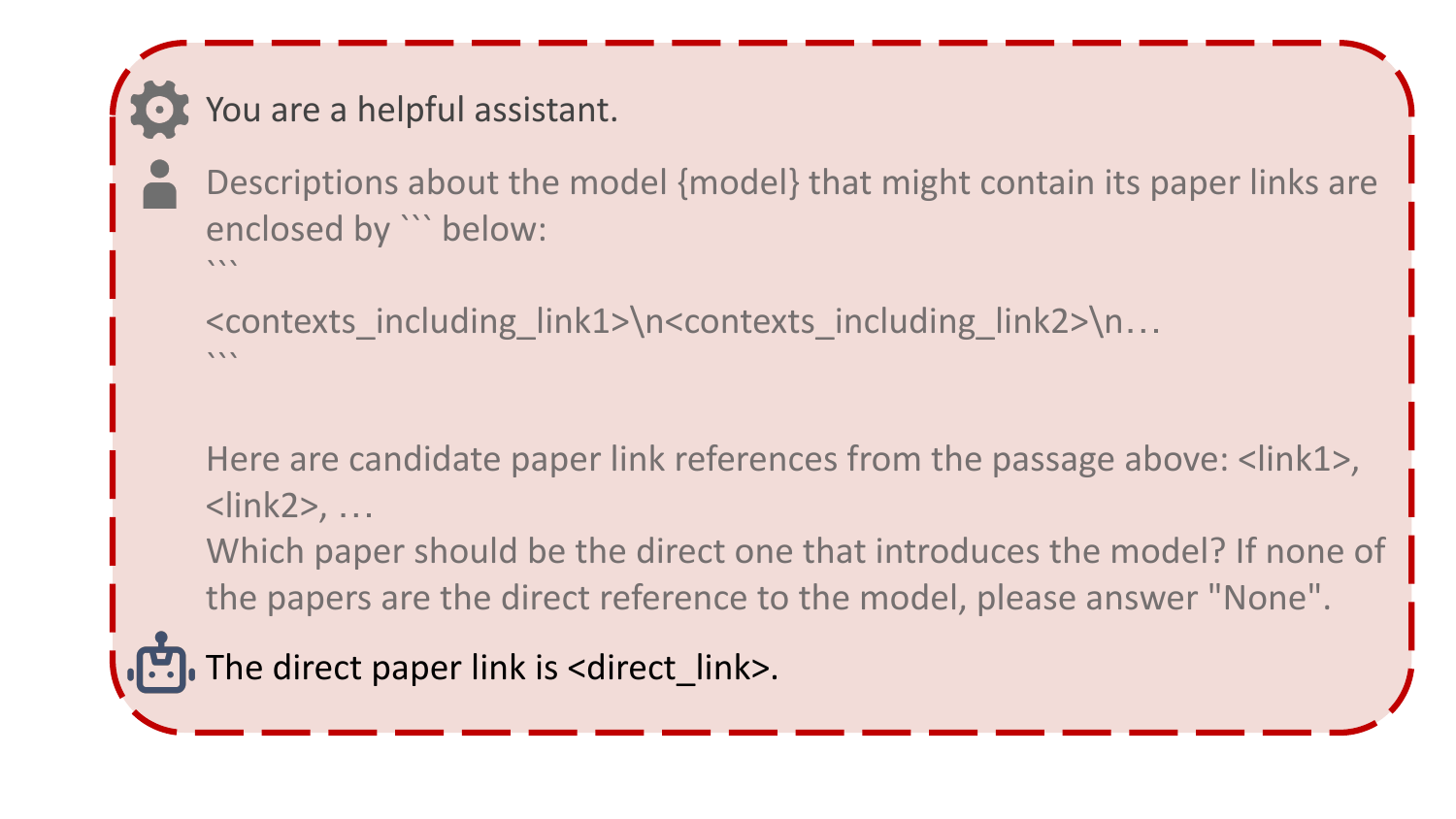}
    \caption{Prompts for calling GPT3.5 to select direct paper links. We prepend one positive example and one negative example to the message list to improve its inference quality.}
    \label{fig:gpt_annotation}
\end{figure}

\paragraph{GPT Annotation Details} We show our two-shot prompts for asking \texttt{GPT-3.5-turbo} to select direct paper links in \cref{fig:gpt_annotation}. Direct GitHub link selection is prompted similarly.

\begin{table}[H]
    \centering
    \resizebox{\columnwidth}{!}{
        \begin{tabular}{lrrrr}
        \toprule
        LLM & \# words & \# sentences & \# links \\
        \hline
        GPT3.5 & 4023.88 & 215.17 & 4.18 \\
        Llama2 70B Chat & 6210.32 & 323.56 & 4.55 \\
        Llama2 7B Chat & 5548.50 & 302.73 & 1.44 \\
        Mistral 7B Inst & 4126.07 & 202.16 & 2.65 \\
        \bottomrule
        \end{tabular}
    }
    \caption{Statistics about whole generated model cards}
    \label{tab:gen_stats}
\end{table}

\section{Dataset Analysis}
\label{appn:model_card_task_taxonomy}

We provide the number of card examples with direct paper links in their human-generated cards, with direct GitHub repository links, and with both links in \cref{tab:stat_card}. We also provide additional figures about the dataset task taxonomy in \cref{fig:task_taxonomy_models,fig:task_taxonomy_models}. The taxonomy is obtained using the REST API of HF Hub.

\begin{table}[H]
    \centering
    \resizebox{\columnwidth}{!}{
    \begin{tabular}{lclcccc}
    \toprule
     & Split & Measure & \# w/ papers & \# w/ repos & \# w/ both \\
     \hline
    \multirow{4}{*}{ModelCard} & \multirow{2}{*}{all} & \# samples & 5689 & 4829 & 2485 \\
                                &  & \# words & 1064 & 948 & 1134 \\ \cline{2-6}
                                & \multirow{2}{*}{test} & \# samples & 344 & 299 & 294 \\
                                &  & \# words & 668 & 710 & 711 \\
    \hline
    \multirow{4}{*}{DataCard} & \multirow{2}{*}{all} & \# samples & 660 & 533 & 328 \\
                                &  & \# words & 1394 & 1104 & 1416 \\ \cline{2-6}
                                & \multirow{2}{*}{test} & \# samples & 86 & 71 & 50 \\
                                &  & \# words & 1003 & 1290 & 1155 \\
    \bottomrule
    \end{tabular}
    }
    \caption{Statistics for crawled model cards and data cards, including the number of examples with direct paper links or direct github links or both, and the average number of words in each category.}
    \label{tab:stat_card}
\end{table}

\begin{figure*}[t]
\minipage{0.5\textwidth}%
  \includegraphics[width=\columnwidth]{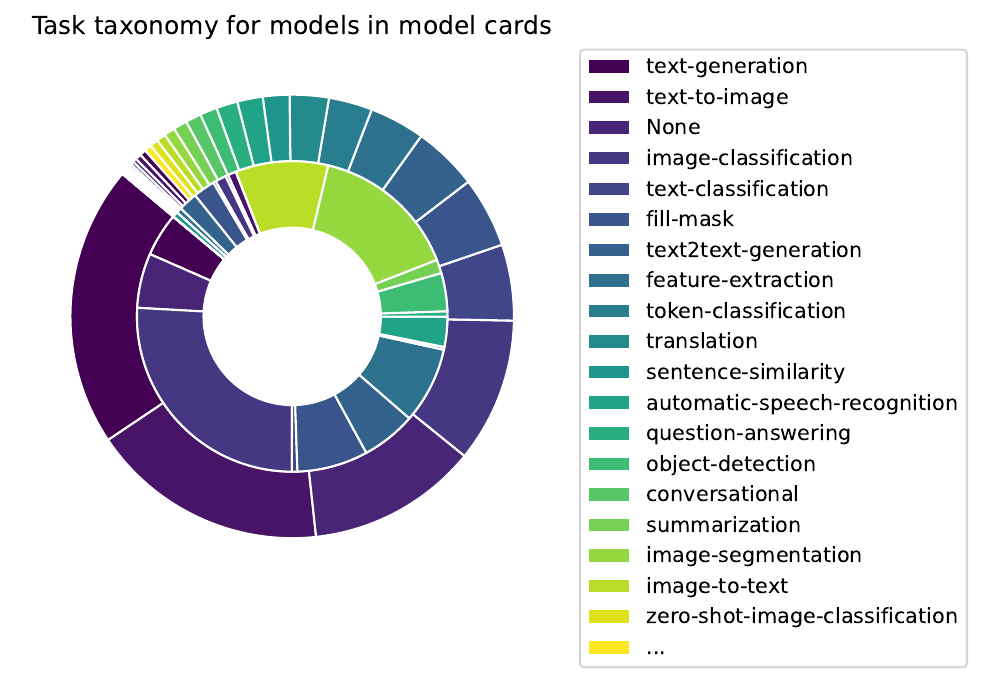}
\endminipage\hfill
\minipage{0.5\textwidth}
  \includegraphics[width=\linewidth]{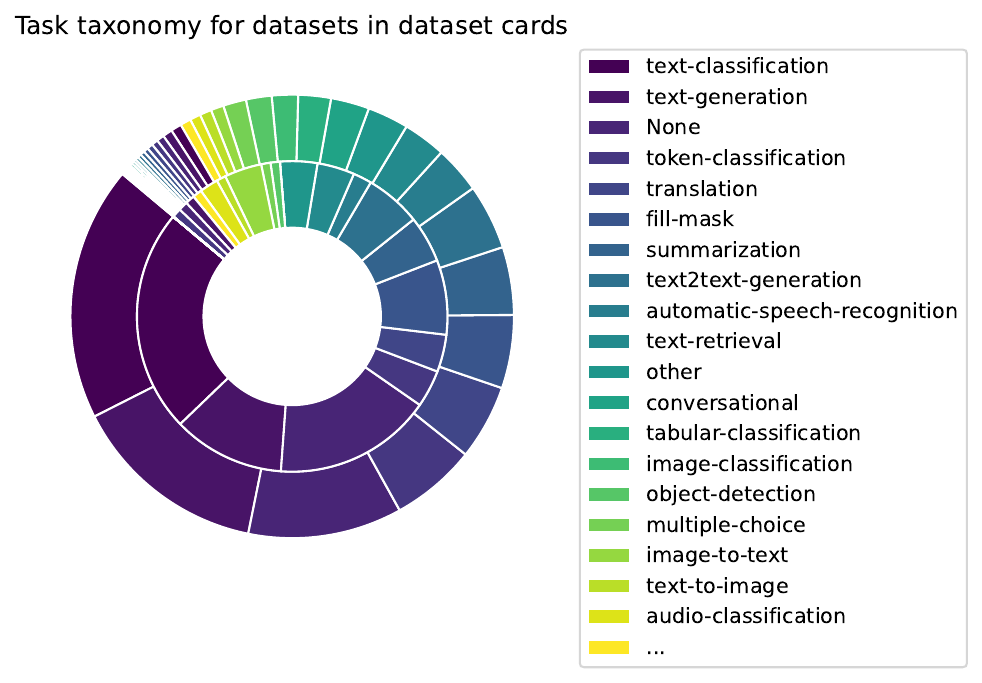}
\endminipage
\caption{The task taxonomy of models in the model cards dataset (left), and the task taxonomy of datasets in the dataset cards dataset (right), with the inner circle as the test set, and the outer circle as the whole set.}
\label{fig:task_taxonomy}
\end{figure*}



\begin{figure}[t]
    \centering
    \includegraphics[width=\columnwidth]{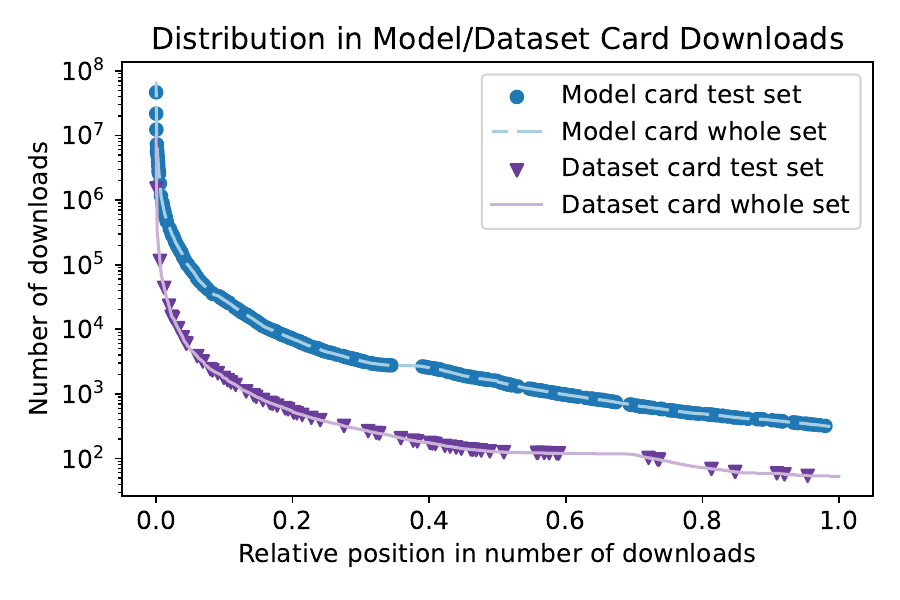}
    \caption{Distribution of the amount of downloads for the whole dataset and the test set. Test set examples distribute quite uniformly.}
    \label{fig:stats_downloads}
\end{figure}

\section{Retriever Details}
\label{appn:retriever}

We use FAISS as our embedding store database \citep{johnson2019billion}. We fix the chunk size as 512 and the chunk overlap as 64. After retrieving relevant sections, we choose to obtain 8 chunks from these sections, together with 4 other chunks from other sections to reduce the bias propagation.

\section{Generator Details}
\label{appn:generator}

\begin{figure*}[t]
    \centering
    \includegraphics[width=\textwidth]{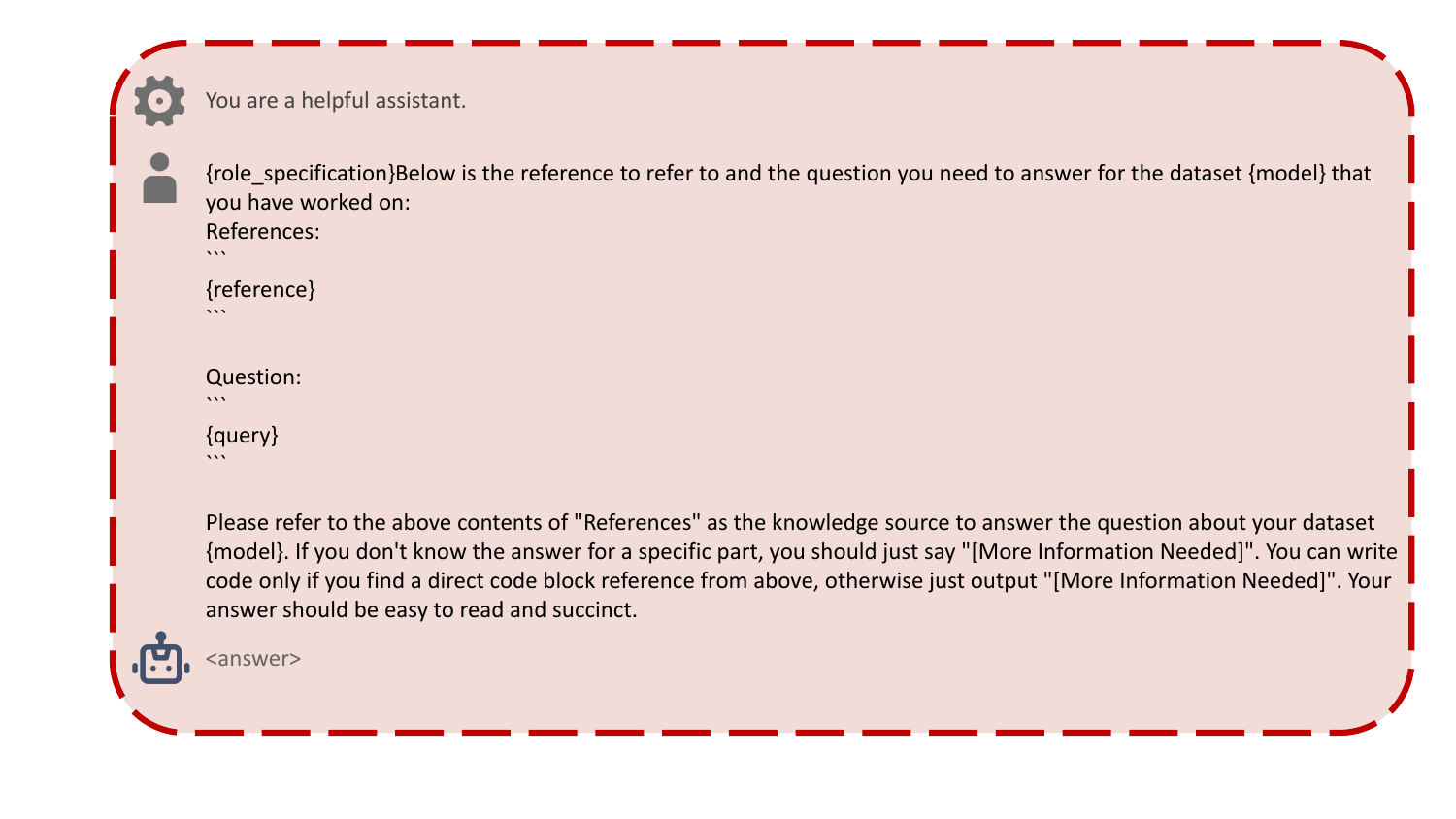}
    \caption{Our generation prompt templates.}
    \label{fig:generation_prompt}
\end{figure*}

Open-sourced LLMs are inferenced through \texttt{vllm} \citet{kwon2023efficient}. \texttt{Llama2-70B-Chat-HF} is run on 4 A6000s. Two 7B models are run on 1 A6000. We fix temperature to 0 to ensure a stable generation quality. We show our prompt description of different roles in \cref{tab:role_specificaion}, and the generation prompt in \cref{fig:generation_prompt}.

\begin{table*}[t!]
    \centering \small
    \begin{tabularx}{\textwidth}{llX}
    \toprule
    Card & Role & Description \\
    \hline
    \multirow{3}{*}{ModelCard} & Developer & who writes the code and runs training \\
    & Sociotechnic & who is skilled at analyzing the interaction of technology and society long-term (this includes lawyers, ethicists, sociologists, or rights advocates) \\
    & Project organizer & who understands the overall scope and reach of the model and can roughly fill out each part of the card, and who serves as a contact person for model card updates \\
    \hline
    \multirow{3}{*}{DataCard} & Data curator & who collects and organizes the data \\
    & Data analyst & who is skilled at understanding and documenting dataset characteristics and biases \\
    & Data manager & who oversees dataset versioning, availability, and usage guidelines \\
    \bottomrule
    \end{tabularx}
    \caption{Our prompts for different roles in answering specific questions.}
    \label{tab:role_specificaion}
\end{table*}

\begin{figure}[h]
    \centering
    \includegraphics[width=\columnwidth]{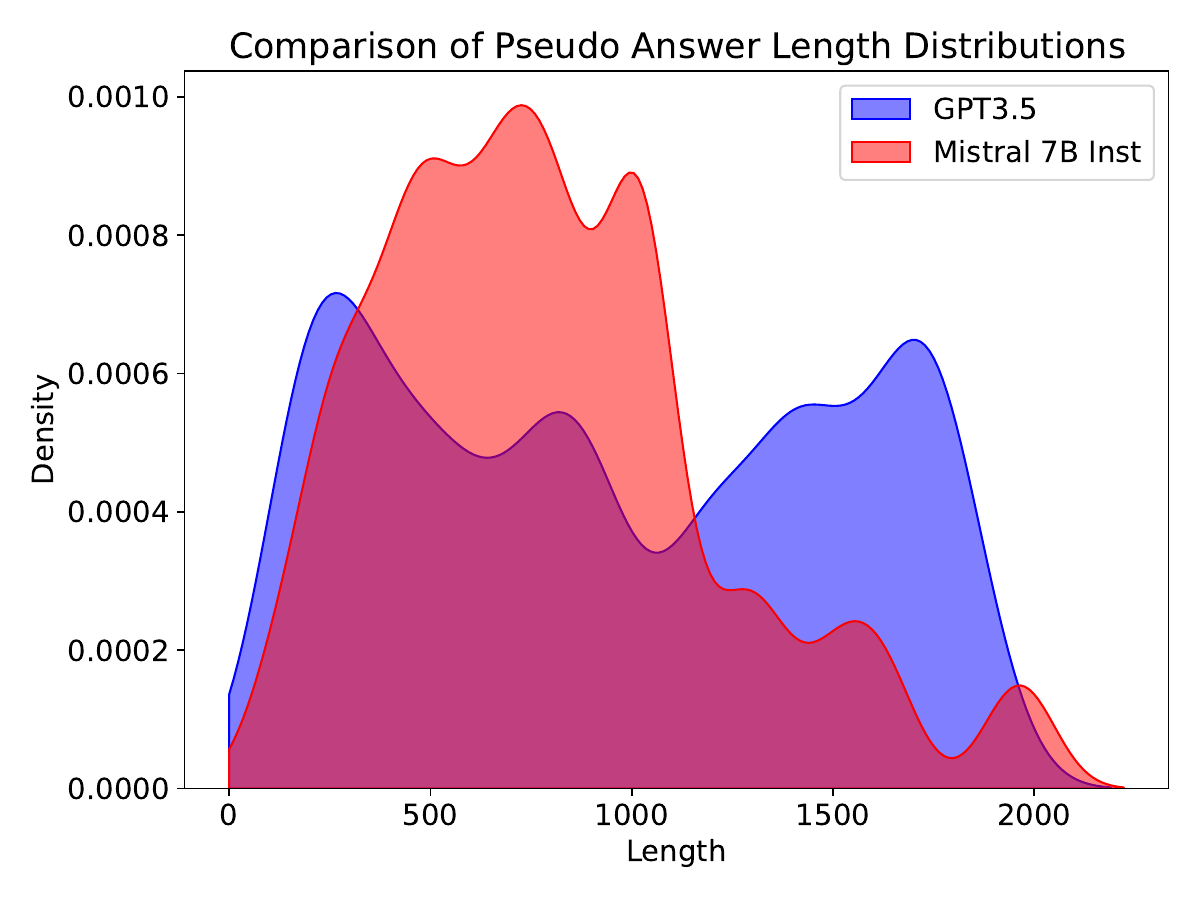}
    \caption{Distribution comparison of pseudo answer length generated by GPT3.5 and Mistral 7B Instruct.}
    \label{fig:pseudo_answer_len}
\end{figure}

\section{LLM Generated Model Card Statistics}
\label{appn:gen_stats}

Statistics about LLM generated model cards are shown in \cref{tab:gen_stats,tab:gen_stats_per_question_words,tab:gen_stats_per_question_sentences,tab:gen_stats_per_question_links}.

\section{Metric Details}
\label{appn:metrics}

For standard metrics, we use the list of retrieved texts together with the generated answer as inputs. We normalize all these scores to be in the [0,1] range. Since the output of \texttt{nli-deberta-v3-large} is in \{``contradiction'', ``entailment'', ``neutral''\}, we map these outputs to \{0, 0.5, 1\}, respectively to maintain a percentage scale. We use the implementation of ROUGE score by HF. We use official implementations for BERTScore and BARTScore.

For GPT metrics, we use \texttt{GPT-4-1106-preview} as evaluators for the main results, and use \texttt{GPT-3.5-turbo} for ablation studies. 

\section{Human Annotation Details}
\label{appn:human_annotate}

\begin{figure*}[t]
    \centering
    \includegraphics[width=\textwidth]{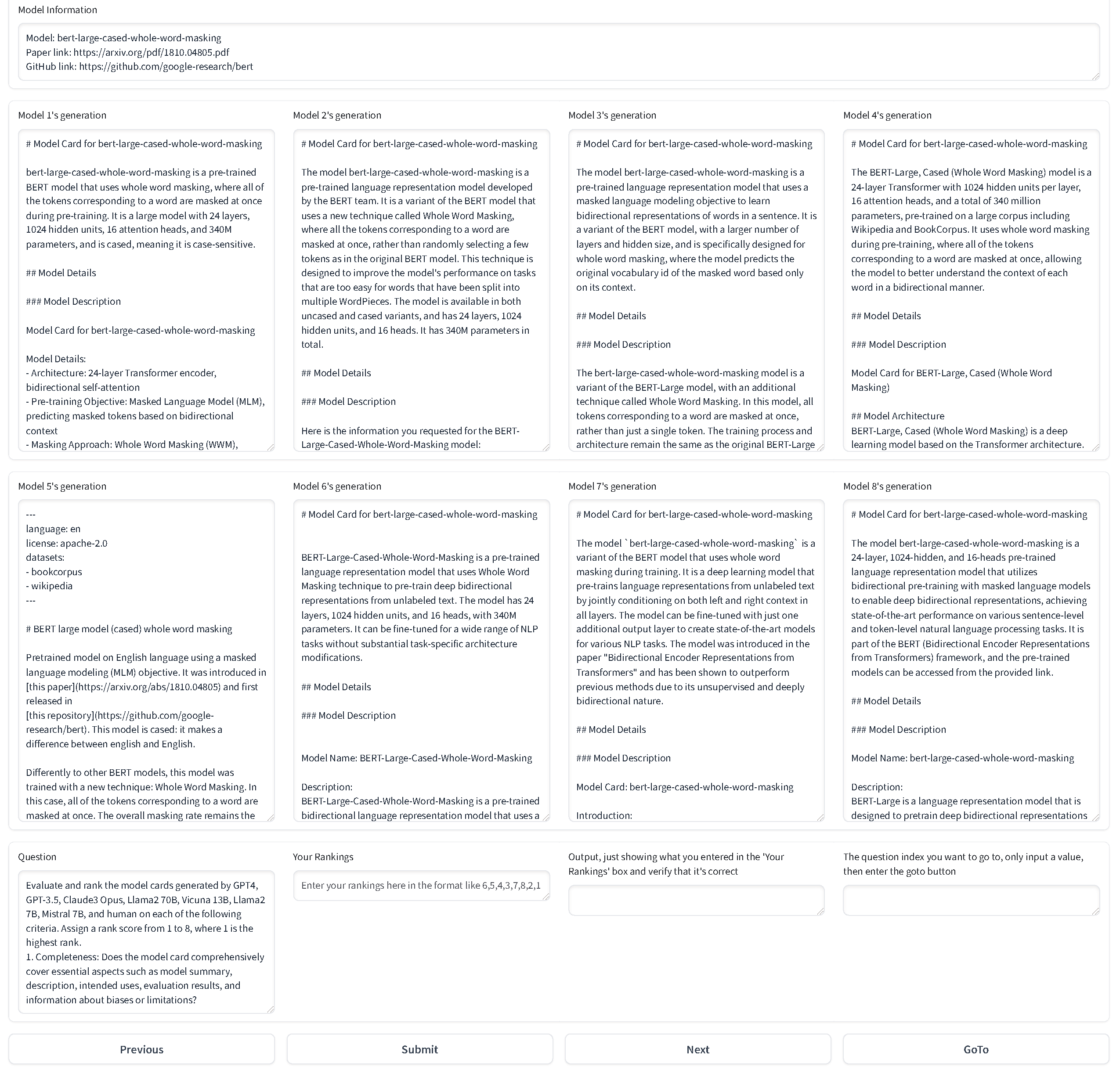}
    \caption{The human annotation interface built by gradio with an example of model \texttt{bert-large-cased-whole-word-masking} \citep{abid2019gradio, devlin2018bert}. The information that a model card is written by whom is hidden, and orders of five model cards shown at each time are randomly shuffled to avoid positional bias.}
    \label{fig:human_interface}
\end{figure*}

We give two annotators the same set of examples each with seven model cards generated by LLMs and one written by human. For each model example for which LLMs generate model cards, we provide annotators with the model name, the corresponding paper link, GitHub link, and a collection of model cards created by humans or LLMs, as illustrated in \cref{fig:human_interface}. We also provide the question template set in \cref{tab:qt_model_card}, along with the following instructions:

Annotators are asked to rank the model cards based on five criteria: completeness, accuracy, objectivity, understandability, and reference quality. The ranking is asked to consider the summation of the binary classification score of whether each question from the model card’s question template is satisfactorily answered according to the specific metric. The final score reported in \cref{tab:human_eval} is calculated by simply subtracting the rank from (1 + the total number of candidates). Further, we define each metric as follows:

\begin{itemize}
    \item Completeness: Does the model card comprehensively cover essential aspects such as model summary, description, intended uses, evaluation results, and information about biases or limitations?
    \item Accuracy: Are answers to all the questions in the model card consistent and accurate compared to the details provided in the model's official paper and GitHub READMEs?
    \item Objectivity: Does the model card present a balanced perspective of the model, recognizing both its strengths and weaknesses?
    \item Understandability: Is the information in the model card clear and easily understandable for both technical and non-technical audiences? Are complex technical concepts explained in a manner that can be easily grasped by users without in-depth technical knowledge?
    \item Reference Quality: Does the model card include necessary citations and references to related papers and links? Do all provided links redirect correctly to their intended URLs?
\end{itemize}

In cases where the summation scores for a question are tied for multiple models, we allow annotators the discretion to rank based on the quality of answers to the most important questions, including model summary, description, intended uses, evaluation results, and information about biases or limitations.

We calculate the Krippendorff's $\alpha$ among the results of two annotators, and got mean=0.68 and std=0.29 for the agreement level. We report averaged ranking scores in \cref{tab:human_eval}. Note that we don't have direct comparison across human evaluation metrics vs. automatic metrics, since our human metrics evaluate on a whole model card, while automatic metrics take each $ (\bm{Q}, \bm{R}, \bm{A})$ tuple for evaluation and they have different scales. We need to implement human metrics in this way to supplement the limited scope of automatic metrics' focus.

\section{Retrieval Only Baseline Details}
\label{appn:baseline2}

\begin{figure*}[t]
    \centering
    \includegraphics[width=\textwidth]{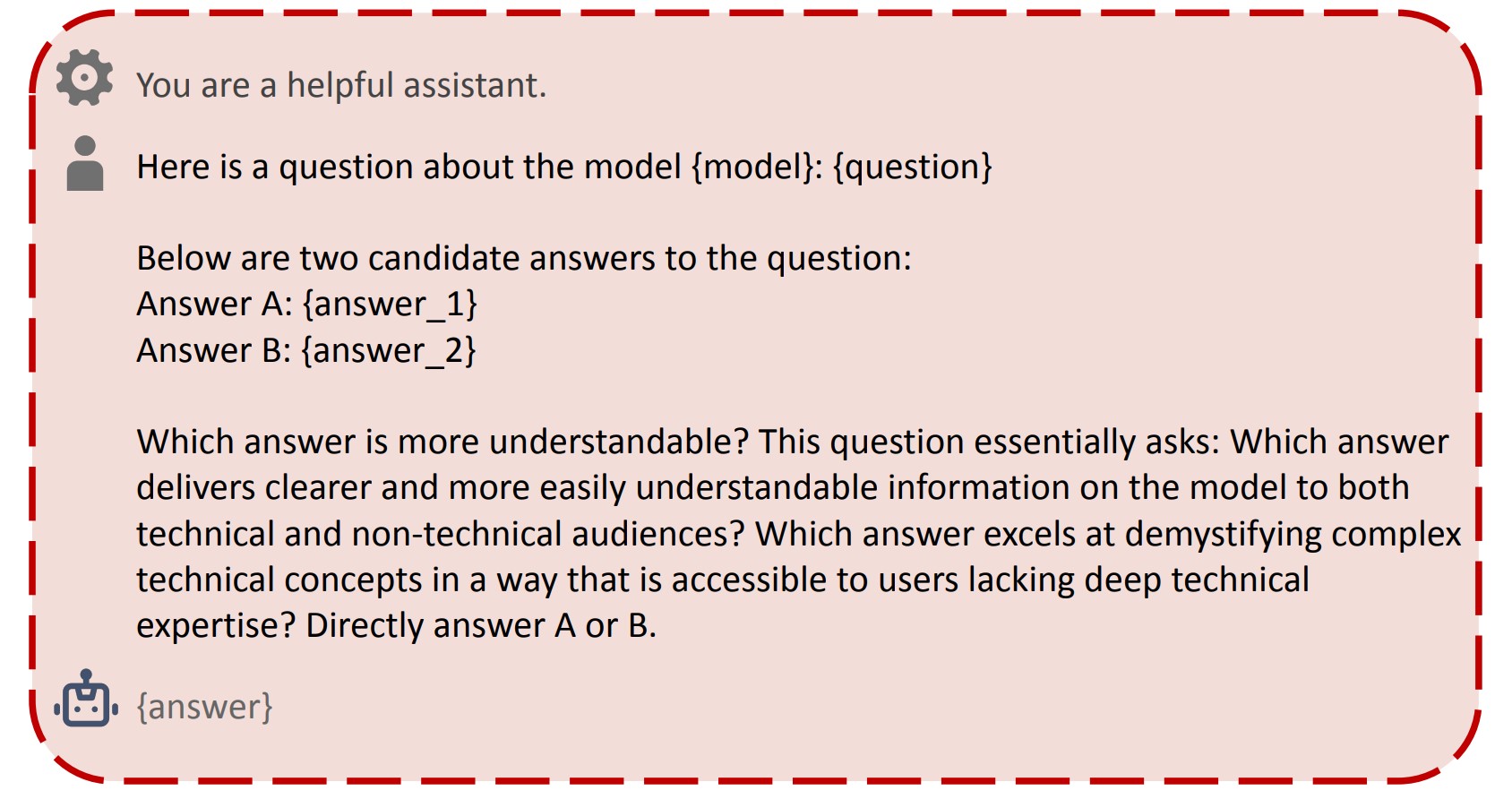}
    \caption{Prompt template to compare \ourmethod's understandability to the retrieval-only baseline.}
    \label{fig:baseline2_prompt_understandability}
\end{figure*}

\begin{figure*}[t]
    \centering
    \includegraphics[width=\textwidth]{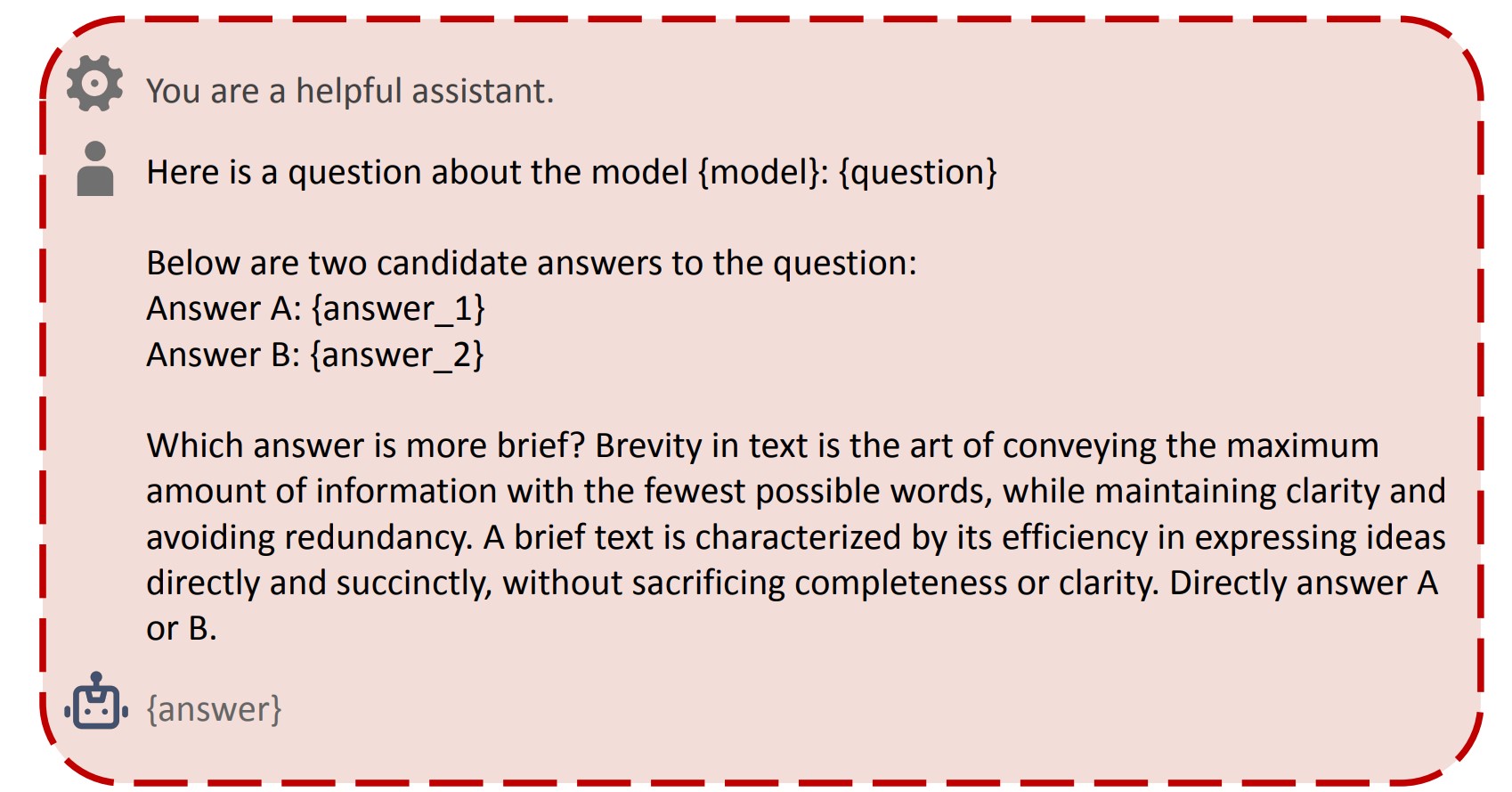}
    \caption{Prompt template to compare \ourmethod's understandability to the retrieval-only baseline.}
    \label{fig:baseline2_prompt_brevity}
\end{figure*}

Following \citet{fu2023gptscore}, we prompt GPT3.5 to assess the understandability and brevity of generated texts according to input questions. Since there are only two methods to evaluate: retrieval-only and \ourmethod, we use the comparative assessment proposed by \citet{liusie-etal-2024-llm}, to compare these two candidates in a pairwise manner. The definition of understandability is the same as in the human annotation. \cref{fig:baseline2_prompt_understandability,fig:baseline2_prompt_brevity} shows the prompt templates we use to generate comparative results. The order of two candidates in the prompt is randomly shuffled to avoid positional bias. We report the ratio of one method better than another for understandability in \cref{tab:baseline2}. Full results including brevity are reported in \cref{tab:baseline2_full}.

\begin{table*}[ht]
    \centering
    \begin{tabular}{l|l|llll}
    \toprule
    \textbf{Model} & \textbf{Method} & \textbf{\# Words} & \textbf{AR} & \textbf{Understandability} & \textbf{Brevity} \\
    \hline
    \multirow{2}{*}{GPT3.5} & Retrieval only & 613.95 & 81.28 & 5.60\% & 1.60\% \\
    & \ourmethod & 200.74 & 90.84 (\green{+9.56}) & \textbf{94.40\%} & \textbf{98.40\%} \\
    \hline
    \multirow{2}{*}{Llama2 70B} & Retrieval only & 645.91 & 81.61 & 1.60\% & 3.20\% \\
    & \ourmethod & 230.42 & 90.32 (\green{+8.71}) & \textbf{98.40\%} & \textbf{96.80\%} \\
    \hline
    \multirow{2}{*}{Llama2 7B} & Retrieval only & 603.45 & 81.32 & 4.40\% & 
2.80\% \\
    & \ourmethod & 203.70 & 90.78 (\green{+9.46}) & \textbf{95.60\%} & \textbf{97.20\%} \\
    \hline
    \multirow{2}{*}{Mistral 7B} & Retrieval only & 590.35 & 81.49 & 2.40\% & 2.40\%\\
    & \ourmethod & 189.11 & 89.83 (\green{+8.34}) & \textbf{97.60\%} & \textbf{97.60\%} \\
    \bottomrule
    \end{tabular}
    \caption{GPT3.5 evaluation results of the retrieval-only baseline and \ourmethod on word numbers, answer relevance, understandability, brevity.}
    \label{tab:baseline2_full}
\end{table*}

\section{Pseudo Answer Ablation Study Analyses}
\label{appn:pseudo_answer_len}

We show the distribution of pseudo answer length generated by GPT3.5 and Mistral 7B Instruct in \cref{fig:pseudo_answer_len}.

\section{Generation Only Ablation Study Analyses}
\label{appn:generation_only}

\begin{figure}[h]
    \centering
    \includegraphics[width=0.9\columnwidth]{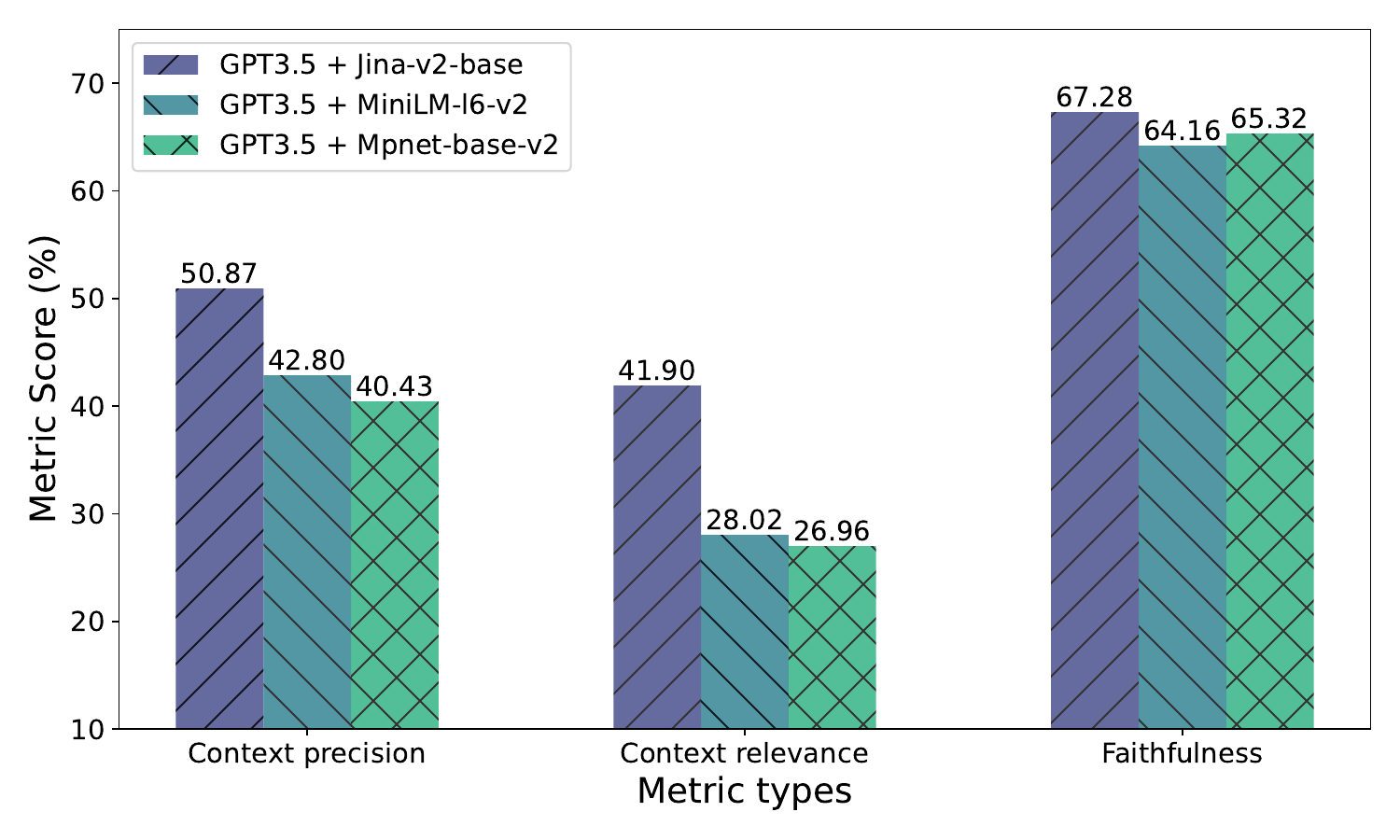}
    \caption{Comparison of three embedding models on context precision, context relevance, and faithfulness.}
    \label{fig:ablation_embed}
\end{figure}

\begin{table}[t]
    \centering
    \resizebox{\columnwidth}{!}{
    \begin{tabular}{lllll}
    \toprule
    Question & GPT3.5 & Llama2 70B & Llama2 7B & Mistral 7B \\
    \hline
    Summary & 53.91 & 89.40 & 71.93 & 63.61 \\
    Description & 275.47 & 276.50 & 187.40 & 264.11 \\
    Funded by & 78.29 & 96.10 & 91.97 & 31.15 \\
    Shared by & 33.41 & 108.62 & 57.94 & 43.69 \\
    Model type & 46.11 & 115.77 & 67.69 & 56.07 \\
    Language & 30.24 & 100.23 & 57.52 & 21.67 \\
    License & 47.56 & 94.86 & 43.05 & 42.63 \\
    Finetuned from & 93.95 & 137.65 & 115.16 & 65.91 \\
    Demo sources & 76.70 & 150.54 & 228.09 & 141.35 \\
    Direct use & 227.26 & 247.95 & 260.14 & 211.97 \\
    Downstream use & 287.05 & 256.03 & 301.56 & 254.17 \\
    Out of scope use & 305.64 & 341.98 & 339.81 & 225.52 \\
    Bias risks limitations & 305.09 & 330.94 & 317.83 & 274.26 \\
    Bias recommendations & 298.46 & 333.96 & 336.44 & 309.82 \\
    Training data & 61.17 & 103.41 & 72.18 & 85.98 \\
    Preprocessing & 169.67 & 285.66 & 228.65 & 222.67 \\
    Training regime & 110.71 & 208.14 & 162.46 & 179.76 \\
    Speeds sizes times & 170.33 & 250.69 & 211.52 & 192.81 \\
    Testing data & 112.20 & 144.15 & 87.29 & 87.16 \\
    Testing factors & 230.03 & 293.02 & 344.08 & 245.14 \\
    Testing metrics & 64.45 & 267.89 & 226.08 & 137.77 \\
    Results & 137.94 & 276.72 & 263.82 & 210.40 \\
    Results summary & 154.57 & 230.82 & 215.33 & 136.51 \\
    Model examination & 214.29 & 317.01 & 264.26 & 169.52 \\
    Hardware & 24.87 & 81.48 & 72.26 & 21.44 \\
    Software & 64.71 & 91.29 & 49.32 & 23.53 \\
    Hours used & 27.95 & 172.74 & 164.28 & 58.86 \\
    Cloud provider & 26.13 & 82.82 & 56.88 & 18.55 \\
    Co2 emitted & 36.01 & 220.29 & 243.23 & 33.65 \\
    Model specs & 207.91 & 276.66 & 204.47 & 161.47 \\
    Compute infrastructure & 51.80 & 227.01 & 205.86 & 134.92 \\
    \bottomrule
    \end{tabular}
    }
    \caption{Number of words in generated model cards per question averaged by all samples in the test set.}
    \label{tab:gen_stats_per_question_words}
\end{table}

\begin{table}[t]
    \centering
    \resizebox{\columnwidth}{!}{
\begin{tabular}{lllll}
    \toprule
    Question & GPT3.5 & Llama2 70B & Llama2 7B & Mistral 7B \\
    \hline
    Summary & 1.95 & 3.23 & 2.61 & 2.39 \\
    Description & 14.51 & 13.87 & 8.93 & 12.87 \\
    Funded by & 4.25 & 4.96 & 6.40 & 1.89 \\
    Shared by & 1.86 & 4.53 & 3.18 & 2.41 \\
    Model type & 1.51 & 3.47 & 2.68 & 1.70 \\
    Language & 1.10 & 4.30 & 1.84 & 1.09 \\
    License & 2.78 & 4.74 & 2.79 & 2.49 \\
    Finetuned from & 4.81 & 5.96 & 5.98 & 3.47 \\
    Demo sources & 3.83 & 7.42 & 12.81 & 6.48 \\
    Direct use & 8.78 & 7.45 & 12.20 & 6.29 \\
    Downstream use & 10.23 & 8.11 & 16.29 & 7.30 \\
    Out of scope use & 16.50 & 21.69 & 20.71 & 10.20 \\
    Bias risks limitations & 19.07 & 22.76 & 19.05 & 16.36 \\
    Bias recommendations & 18.04 & 22.13 & 19.88 & 18.44 \\
    Training data & 3.14 & 4.54 & 3.31 & 4.01 \\
    Preprocessing & 11.06 & 18.20 & 13.34 & 12.46 \\
    Training regime & 4.82 & 12.66 & 7.19 & 11.08 \\
    Speeds sizes times & 8.41 & 12.74 & 10.62 & 9.40 \\
    Testing data & 7.98 & 9.00 & 5.55 & 4.96 \\
    Testing factors & 13.26 & 17.23 & 21.64 & 11.60 \\
    Testing metrics & 3.67 & 14.11 & 14.20 & 7.12 \\
    Results & 7.69 & 16.85 & 16.22 & 10.50 \\
    Results summary & 9.01 & 10.94 & 9.79 & 6.21 \\
    Model examination & 11.32 & 17.74 & 15.67 & 8.47 \\
    Hardware & 1.73 & 4.29 & 3.43 & 1.39 \\
    Software & 3.50 & 4.54 & 2.45 & 1.47 \\
    Hours used & 2.06 & 7.52 & 8.29 & 2.86 \\
    Cloud provider & 1.82 & 4.38 & 2.92 & 1.32 \\
    Co2 emitted & 2.40 & 9.14 & 10.27 & 2.13 \\
    Model specs & 10.52 & 12.12 & 9.90 & 7.17 \\
    Compute infrastructure & 3.59 & 12.94 & 12.61 & 6.61 \\
    \bottomrule
    \end{tabular}
    }
    \caption{Number of sentences in generated model cards per question averaged by all samples in the test set.}
    \label{tab:gen_stats_per_question_sentences}
\end{table}

\begin{table}[t]
    \centering
    \resizebox{\columnwidth}{!}{
\begin{tabular}{lllll}
\toprule
Question & GPT3.5 & Llama2 70B & Llama2 7B & Mistral 7B \\
\hline
Summary & 0.02 & 0.05 & 0.00 & 0.01 \\
Description & 0.17 & 0.04 & 0.01 & 0.04 \\
Funded by & 0.37 & 0.06 & 0.05 & 0.06 \\
Shared by & 0.36 & 0.58 & 0.04 & 0.12 \\
Model type & 0.00 & 0.00 & 0.00 & 0.00 \\
Language & 0.01 & 0.00 & 0.00 & 0.01 \\
License & 0.53 & 0.82 & 0.17 & 0.36 \\
Finetuned from & 0.26 & 1.06 & 0.30 & 0.49 \\
Demo sources & 0.66 & 0.82 & 0.51 & 0.94 \\
Direct use & 0.34 & 0.05 & 0.01 & 0.09 \\
Downstream use & 0.17 & 0.03 & 0.02 & 0.04 \\
Out of scope use & 0.20 & 0.00 & 0.00 & 0.00 \\
Bias risks limitations & 0.01 & 0.00 & 0.00 & 0.00 \\
Bias recommendations & 0.04 & 0.01 & 0.00 & 0.00 \\
Training data & 0.29 & 0.24 & 0.00 & 0.02 \\
Preprocessing & 0.04 & 0.03 & 0.00 & 0.01 \\
Training regime & 0.00 & 0.03 & 0.01 & 0.01 \\
Speeds sizes times & 0.21 & 0.10 & 0.02 & 0.05 \\
Testing data & 0.01 & 0.01 & 0.03 & 0.02 \\
Testing factors & 0.01 & 0.00 & 0.00 & 0.01 \\
Testing metrics & 0.01 & 0.02 & 0.00 & 0.01 \\
Results & 0.03 & 0.05 & 0.03 & 0.04 \\
Results summary & 0.04 & 0.03 & 0.04 & 0.09 \\
Model examination & 0.19 & 0.04 & 0.02 & 0.02 \\
Hardware & 0.00 & 0.02 & 0.04 & 0.01 \\
Software & 0.03 & 0.12 & 0.00 & 0.04 \\
Hours used & 0.01 & 0.02 & 0.00 & 0.01 \\
Cloud provider & 0.03 & 0.11 & 0.04 & 0.02 \\
Co2 emitted & 0.01 & 0.11 & 0.00 & 0.00 \\
Model specs & 0.11 & 0.04 & 0.01 & 0.03 \\
Compute infrastructure & 0.02 & 0.05 & 0.05 & 0.10 \\
\bottomrule
\end{tabular}
    }
    \caption{Number of links in generated model cards per question averaged by all samples in the test set.}
    \label{tab:gen_stats_per_question_links}
\end{table}

\end{document}